\documentclass[journal]{IEEEtran}
\usepackage{amsmath,amsfonts}
\usepackage{algorithmic}
\usepackage{algorithm}
\usepackage{array}
\usepackage{textcomp}
\usepackage{stfloats}
\usepackage{url}
\usepackage{verbatim}
\usepackage{graphicx}
\usepackage{cite}
\usepackage{xcolor}
\usepackage{adjustbox}
\usepackage{subcaption}
\usepackage{multirow}
\usepackage{makecell} 
\usepackage{amssymb}
\usepackage[british]{babel}

\begin{document}

\title{Optimising Event-Driven Spiking Neural Network with Regularisation and Cutoff}


\author{Dengyu Wu, Gaojie Jin, Han Yu, Xinping Yi and Xiaowei Huang 
\thanks{D. Wu and X. Huang are with Department of Computer Science, University of Liverpool, Liverpool, United Kingdom. (E-mails: dengyu.wu@liverpool.ac.uk, xiaowei.huang@liverpool.ac.uk) X. Yi is with National Mobile Communications Research Laboratory, Southeast University, Nanjing, China. G. Jin is with State Key Laboratory of Computer Science, Institute of Software, CAS Beijing, China. H. Yu is with Department of Electrical Engineering, Chalmers University of Technology, Gothenburg, Sweden. \includegraphics[height=8pt, width=8pt]{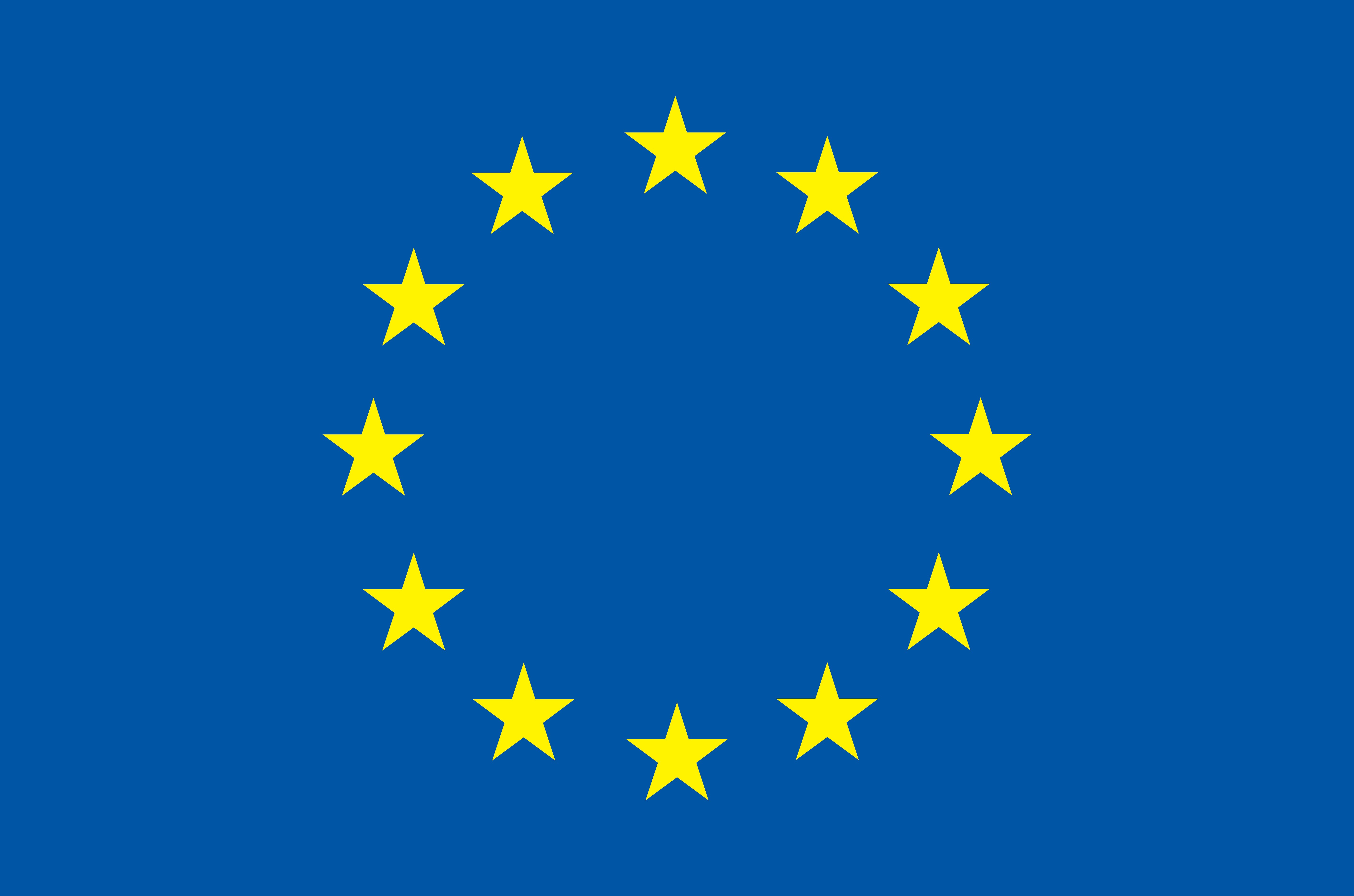} This project has received funding from the European Union’s Horizon 2020 research and innovation programme under grant agreement No 956123, and is also supported by the UK EPSRC under project [EP/T026995/1].}%
}




\maketitle

\begin{abstract}
Spiking neural network (SNN), as the next generation of artificial neural network (ANN), offer a closer mimicry of natural neural networks and hold promise for significant improvements in computational efficiency. However, the current SNN is trained to infer over a fixed duration, overlooking the potential of dynamic inference in SNN. In this paper, we strengthen the marriage between SNN and event-driven processing with a proposal to consider a cutoff in SNN, which can terminate SNN anytime during inference to achieve efficient inference. Two novel optimisation techniques are presented to achieve inference efficient SNN: a Top-K cutoff and a regularisation. The proposed regularisation influences the training process, optimising SNN for the cutoff, while the Top-K cutoff technique optimises the inference phase. We conduct an extensive set of experiments on multiple benchmark frame-based datasets, such as CIFAR10/100, Tiny-ImageNet, and event-based datasets, including CIFAR10-DVS, N-Caltech101 and DVS128 Gesture. The experimental results demonstrate the effectiveness of our techniques in both ANN-to-SNN conversion and direct training, enabling SNNs to require 1.76 to 2.76 $\times$ fewer timesteps for CIFAR-10, while achieving 1.64 to 1.95$\times$ fewer timesteps across all event-based datasets, with near-zero accuracy loss. These findings affirms the compatibility and potential benefits of our techniques in enhancing accuracy and reducing inference latency when integrated with existing methods. Code available: https://github.com/Dengyu-Wu/SNNCutoff
\end{abstract}


\section{Introduction} \label{sec:introduction}

Spiking neural network (SNN) has recently attracted significant research and industrial interest thanks to its energy efficiency and low latency, and there are neuromorphic chips such as Loihi \cite{davies2018loihi} and TrueNorth \cite{akopyan2015truenorth} on which SNN can be deployed. Mechanistically, SNN mimics biological neurons, which process and forward spikes independently. With such an asynchronous working mechanism, only a small subset of neurons will be activated during inference. In essence, SNN is inherently efficient in terms of computation. 

\begin{figure*}[!ht]
    \centering
    \includegraphics[width=0.88\textwidth]{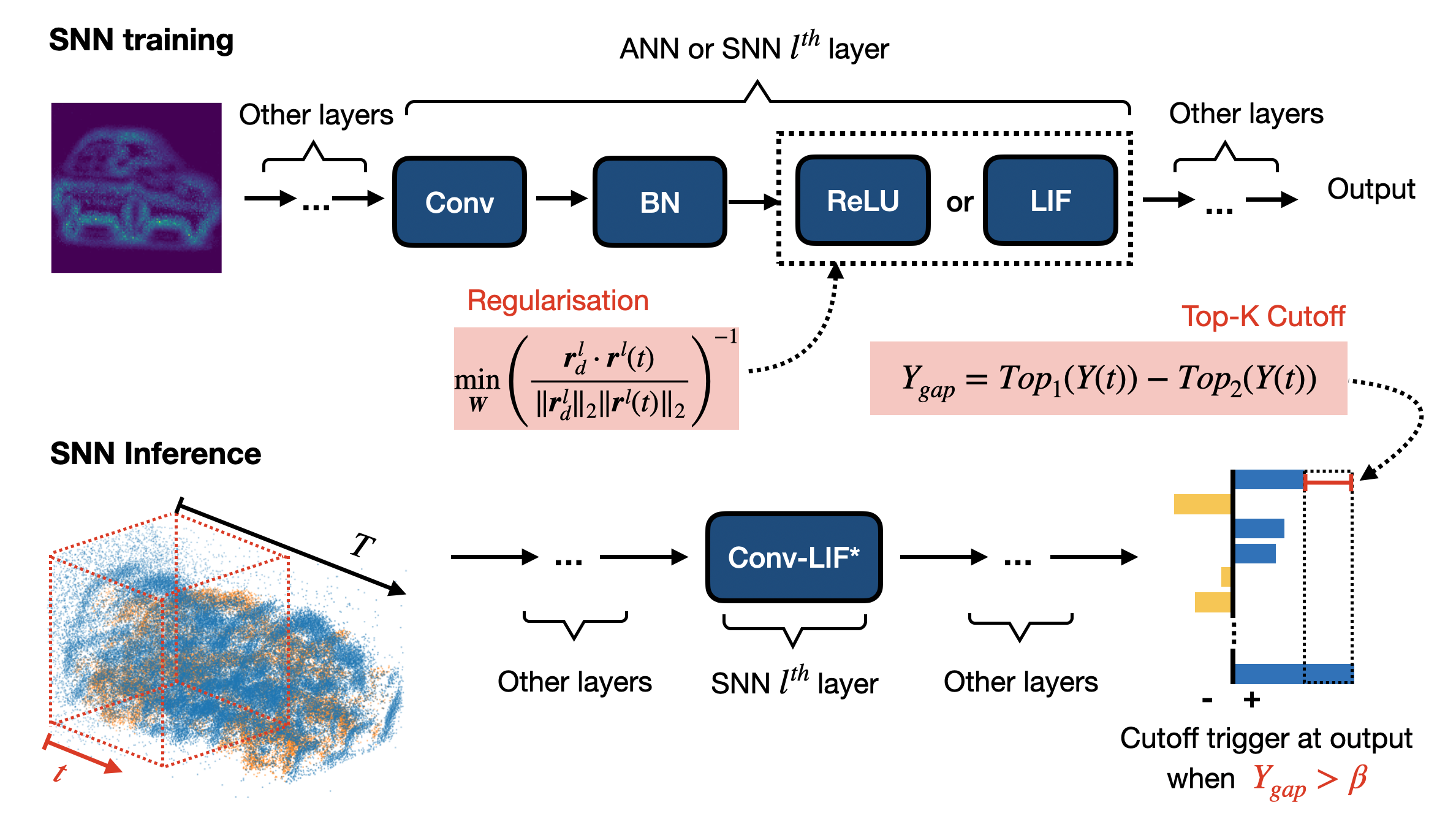}
    \caption{An illustrative diagram showing the regularisation for optimising SNN and the cutoff mechanism for reducing latency on CIFAR10-DVS dataset. Cutoff is triggered when $Y_{\text{gap}}$ is greater than $\beta$, a value dynamically determined by a confidence rate as introduced in Section \ref{sec:cutoffmech}. }
	\label{fig:introduction}

\end{figure*}

The asynchronous mechanism indicates that SNN could be more effective when used with event-based input. Neuromorphic sensors such as dynamic vision sensor (DVS) \cite{DVS:2008,DVS:2010,DVS:2020} and dynamic audio sensor (DAS) \cite{DAS:2018} have been developed to generate binary ``events'', which are ideal inputs to SNN. For instance, unlike conventional frame-based cameras, which measure the ``absolute'' brightness at a constant rate, DVS cameras are bio-inspired sensors that \emph{asynchronously} measure per-pixel 
brightness changes, and output a stream of events that encode the time, location, and sign of the brightness changes \cite{DVS:2020}. 
DVS adapts dynamically to the scene's activity, with fewer events in static scenes and a higher volume when significant changes occur. Consequently, the energy and bandwidth consumption scale efficiently based on actual demand \cite{amir2017low,kim2021n}, and leads to SNN operates at sparse manner \cite{messikommer2020event}.
However, an additional encoding step, such as rate-based coding \cite{rueckauer2017conversion} and Poisson coding \cite{sengupta2019going}, is necessary for frame-based input before forward propagation in SNN.
Irrespective of the input type---event-based or frame-based--- SNN can deliver sequential predictions at their outputs, showcasing that they can predict at any timestep. To harness these features, we explore cutoff optimal SNN, which allows the termination at any time during the inference on a spike train (i.e., an input) and returns the best possible inference result.

One approach to train SNN is through ANN-to-SNN conversion, which leverages the mature training regime of ANN to first train a high-accuracy ANN and then convert it into an SNN. This method has led to research focused on achieving near-zero conversion loss \cite{deng2021optimal,bu2022optimal,han2020rmp}.
Another methodology involves the use of backpropagation in SNN training \cite{wu2018spatio,wu2019direct, fang2021deep, fang2021incorporating}. Due to the non-differentiable nature of spiking, this approach necessitates the deployment of the surrogate gradient \cite{wu2018spatio,wu2019direct}. 
In this paper, we explore a novel cutoff mechanism in SNN and propose general optimisation strategies for this process. Our ultimate goal is to develop an optimal SNN that effectively balances accuracy and latency. 

This paper makes two key technical contributions. 
Firstly, instead of always predicting at the maximum timestep $T$, we explore an early cutoff mechanism that allows the SNN model to achieve optimal latency and computing efficiency automatically.
As shown in Figure~\ref{fig:introduction}, the SNN model runs a monitoring mechanism to determine when it is sufficiently confident to make a decision. Once such a decision is made at timestep $t < T$,  i.e., $t \in \{1,2,\dots,T\}$, a cutoff action is triggered so that the SNN will not take future inputs until $T$. Therefore, this approach leads to lower latency and fewer computations because a decision is made at time $t$ rather than $T$. 

The second contribution is a proposed regularisation technique for improving SNN cutoff performance. This technique influences the activation distribution during ANN or SNN training, resulting in an SNN that can potentially classify with less input information. As will be discussed in Method (Section~\ref{sec:methods}), our proposed regulariser effectively mitigates the impact of `worst-case' inputs during both ANN and SNN training phases. These worst-case samples are typically inputs that can cause failures in the early inference. Experiments in Section~\ref{exp:accexp} show that we can enhance the state-of-the-art methods including ANN-to-SNN conversion and direct training. 

To facilitate the analysis, we use the following notations throughout the paper: \textbf{bold~symbol} represents a vector,  $l$ to denote the layer index, and $i$ to denote the element index. For example, $\boldsymbol{a}^l$ is a vector and $a_{i}^l$ is the $i$-th element in $\boldsymbol{a}^l$. $\boldsymbol{W}^l$ is weight matrix at the $l$-th layer.

\section{Related Work}\label{sec:related}

The implementation of SNN involves two phases: training and inference. %
The training algorithms for SNN can be broadly categorised into two main approaches: ANN-to-SNN conversion and direct training.

\paragraph{ANN-to-SNN Conversion.} ANN-to-SNN conversion is a widely studied approach for converting a pre-trained ANN into an SNN model. This process relies on the average spiking rate of neurons, which is closely linked to the normalised activation of the rectified linear unit (ReLU) function in the ANN \cite{rueckauer2017conversion}. Early studies in ANN-to-SNN conversion, such as \cite{diehl2015fast, rueckauer2017conversion}, utilised the maximum activation value in each layer of the ANN to normalize the corresponding weights. \cite{sengupta2019going} demonstrated an alternative approach, where normalisation can be achieved by greedily searching for the optimal threshold using the input spike train. A unified conversion framework was introduced in \cite{wu2022little}, which incorporates a scaling factor that can be applied to either the threshold or the weights. Additionally, this framework includes the thresholding for residual elimination to mitigate information loss at the last timestep, further enhancing conversion efficiency. Recent work  \cite{deng2021optimal,wu2022little} shows that outlier elimination in ANN activations can be implemented by applying a clipping operation after the ReLU. Based on this, \cite{pmlr-v139-li21d,bu2022optimal} further minimise the quantisation error by quantisation-aware training. Besides, there are other hybrid methods to fine tune the weights in the converted SNN. For example, \cite{Rathi2020Enabling,rathi2021diet} that combine conversion and direct training. Tandem Learning \cite{wu2021progressive} leverages the gradient from ANN to update SNN during training.

\paragraph{Direct Training.} In contrast to the conversion-based approach, direct training of SNN offers the capability to process temporal features effectively \cite{fang2021incorporating, yao2021temporal}. Numerous studies focused on designing surrogate gradients \cite{wu2018spatio, wu2019direct, neftci2019surrogate, li2021differentiable} to tackle the non-differentiable nature of spike generation in SNN, thereby enabling efficient backpropagation. Reference \cite{yao2021temporal} demonstrates how integrating temporal attention can significantly bolster SNN performance. Advancements in temporal batch normalisation \cite{kim2021revisiting, duan2022temporal} have been proposed to normalise the input current to SNN layers, thereby expediting the convergence process during direct training. Meanwhile, \cite{guo2023membrane} introduced an additional membrane potential normalisation, applied after updating the membrane potential with the input current.

Alongside direct training, additional training optimizations, such as knowledge distillation from ANNs \cite{guo2023joint, zhang2024enhancing}, have been explored as complementary methods to enhance SNN representations. Furthermore, spike-based transformers \cite{zhou2023spikformer, zhang2022spiking, wang2023masked} provide new perspectives by adapting the self-attention mechanism for spike-based computation, replacing complex multiplications with efficient spike-based operations using spike-form queries, keys, and values. Transitioning from conventional binary spikes to ternary spikes (including negative spikes) has also demonstrated performance gains with negligible increases in energy cost \cite{guo2024ternary}. Additionally, shrinking the maximum inference timestep as layers deepen effectively reduces average inference latency \cite{ding2024shrinking}.

\paragraph{Adaptive Inference in SNN.} Despite these significant advancements in training algorithms, a fundamental limitation persists: the optimisation of SNN is predominantly focused on specific inference duration, neglecting the potential for adaptive inference. The exploration of adaptive inference in SNN is still in its early stages, with only a limited number of studies. Specifically, \cite{li2023seenn} introduced an additional deep network to trigger the early-exiting in SNN, a solution that may be resource-intensive for small SNN. Similarly, \cite{li2023unleashing} focuses on ANN-to-SNN conversion by applying network calibration before dynamic prediction, while \cite{chen2023agreeing} employs conformal prediction \cite{shafer2008tutorial} as a trigger mechanism. However, these approaches do not address the optimisation of SNN training for such scenarios requiring dynamic timesteps. Our study aims to fill this gap by combining the cutoff with a general regulariser. 

\section{Leaky Integrate-and-Fire Model}\label{sec:ifmodel} 
\begin{figure*}[t!]
    \centering
	\includegraphics[width=0.96\textwidth]{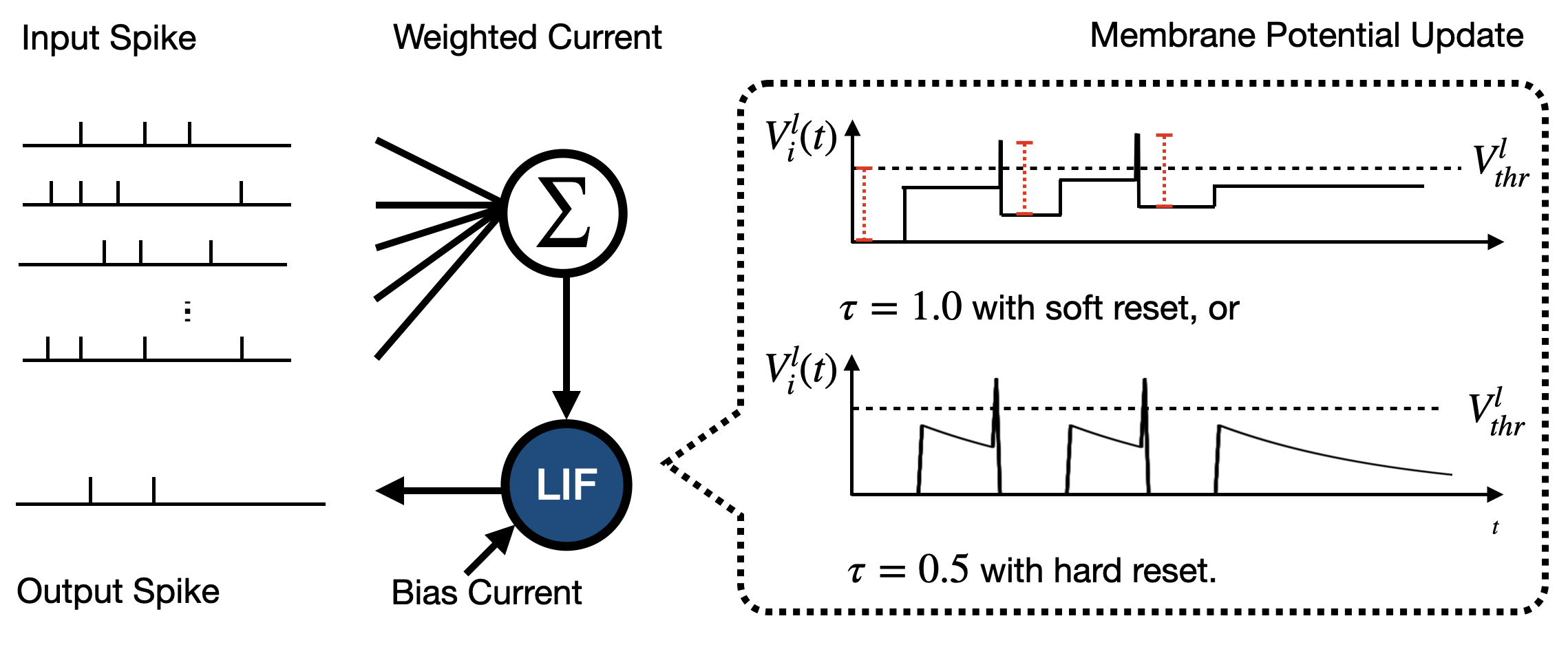}
	\caption{Illustration of inference process in a LIF neuron within the hidden layer. Input spikes charge the membrane potential $V^l_i(t)$ through weighted and bias currents learned during the training stage. When $V^l_i(t)$ reaches the threshold $V^l_{\text{thr}}$, the neuron will generate a spike and then reset the $V^l_i(t)$. In our study, the Integrate-and-Fire (IF) neuron with soft reset is a special case of LIF model when $\tau=1$.} 
	\label{fig:lifmodel}
\end{figure*}

The Leaky Integrate-and-Fire (LIF) model is a foundational component in SNN study, and it is lauded for its simplicity and resemblance to biological neural processing. Figure \ref{fig:lifmodel} illustrates the inference process in a single LIF neuron. The dynamic updates in the LIF neuron are described as

\begin{equation} \label{eq:em_potential}
    \boldsymbol{V}^{l-}(t) = 
    \left\{
    \begin{array}{ll}
        \boldsymbol{V}^{l+}(t-1) + \boldsymbol{Z}^{l}(t), & \text{for } l > 1,  \\
        \boldsymbol{Z}^1(t), & \text{for } l = 1,
    \end{array}
    \right.
\end{equation}
where $\boldsymbol{V}^{l-}$ and $\boldsymbol{V}^{l+}$ denote the vector of membrane potentials before and after reset respectively. The reset process of \(V_i^l(t)\) is categorised as

\begin{equation} \label{eq:reset_process}
    V_i^{l+}(t) = 
    \left\{
    \begin{array}{ll}
       \tau V_i^{l-}(t)(1 - \theta_i^l(t)),  & \text{for hard reset,}  \\
        \tau V_i^{l-}(t) - V^l_{\text{thr}}\theta_i^l(t), & \text{for soft reset,}
    \end{array}
    \right.
\end{equation}
where \(\boldsymbol{\theta}^l(t)\) is a step function, i.e., \(\theta_i^l(t) = 1\) if \(V_i^l(t) \geq V^l_{\text{thr}}\) and \(\theta_i^l(t) = 0\) otherwise. $\tau$ is decay factor and soft reset is introduced for ANN-to-SNN conversion. Specifically, conversion-based SNN inference with Integrate-and-Fire (IF) neuron ($\tau = 1.0$) with soft reset aimed at reducing the information loss caused by the conversion \cite{ruckauer2019closing, han2020rmp, wu2022little}.

In hidden layers, the weighted current $\boldsymbol{Z^l}(t)$ is given by
\begin{align} \label{eq:em_potential1}
    \boldsymbol{Z}^l(t) = \boldsymbol{W}^l\boldsymbol{\theta}^{l-1}(t) + \boldsymbol{b}^l \hspace{0.5cm}\text{ when } l>1,
\end{align}
where $\boldsymbol{W}^l$ is the weight matrix, and $\boldsymbol{b}^l$ is bias current. According to different inputs, $\boldsymbol{Z}^l(t)$ at the first layer, i.e., $\boldsymbol{Z}^1(t)$, can be initialised as either 
\begin{align} \label{eq:ze}
    \textit{Event-based input:} ~~\boldsymbol{Z}^1_e(t) = \boldsymbol{W}^1\boldsymbol{X}(t) + \boldsymbol{b}^1,
\end{align}
where $\boldsymbol{X}(t)$ is the time-dependent spike train, i.e., the input may change the charging current with time during the inference, or
\begin{align} \label{eq:zf}
    \textit{Frame-based input:} ~~\boldsymbol{Z}_f^1 = \boldsymbol{W}^1\boldsymbol{\bar X} + \boldsymbol{b}^1,
\end{align}
where $\boldsymbol{\bar X}$ represents the constant current based on inputs, e.g., normalised pixel value of RGB Image. 
For each timestep, the first layer of SNN transforms the frame-based input $\boldsymbol{\bar X}$ into weighted current $\boldsymbol{Z}_f^1$, which then stimulates the LIF neurons in the first layer to generate spikes.
Notably, for event-based input, SNN can manifest faster inference due to immediate response after receiving the first spike, and it completes the inference whenever the spike train ends, i.e., at $T$. The event-based benchmarks are further introduced in Section \ref{exp:dataset}. This characteristic makes it possible that the inference time is dynamic for different inputs. In this paper, with the cutoff technique as in Section~\ref{sec:cutoffmech}, we will show that the average latency of the inference in SNN can be further reduced (to some $t \leq T$).

\section{Methods}\label{sec:methods}
Section~\ref{sec:cutoffmech} presents the theoretical underpinning of the cutoff mechanism for the inference. Following this, Section~\ref{sec:reg} details the design of a general regulariser to optimise SNN regarding the cutoff mechanism. 

\subsection{Cutoff Mechanism in SNN}\label{sec:cutoffmech}

Thanks to its asynchronous working mechanism, the event-driven SNN can predict when only part of the spike train is processed. However, a naive cutoff on the spike train length (or the event sensor's sampling time) can easily lead to accuracy loss. In this section, we propose a principled method to determine the inference time.

Our approach begins with a theoretical analysis of the cutoff in Section \ref{sec:cutoff-theory}, where we identify the optimal cutoff timestep for each input. Specifically, we explore the optimal timesteps that consistently allow the SNN to make correct predictions in subsequent processing. This evaluation can only be conducted in simulation and requires knowledge of future predictions.

To approximate this process in practice, we introduce the $\text{Top-K}$ cutoff in Section \ref{sec:gap-based}. To elaborate, we define a confidence rate, denoted as $C(t, D\{Y_{\text{gap}} > \beta_t\})$, based on the statistical characteristics of processing a set $D$ of inputs with respect to the discrete time $t$ and the gap between the logits of output neurons, $Y_{\text{gap}}$. The condition {$Y_{\text{gap}} > \beta_t$} is used to identify the samples in $D$ that are suitable for cutoff. We can plot a curve of the confidence rate $C(t, D\{Y_{\text{gap}} > \beta_t\})$ with respect to time $t$ and constant values $\beta_t$. A set of ${\beta_t}$ is extracted from training samples to trigger the Top-K cutoff.

\subsubsection{Optimal Cutoff Timestep} \label{sec:cutoff-theory}
To define a theoretically optimal cutoff within SNN, we propose identifying the cutoff point where subsequent predictions remain positively true. This criterion establishes the cutoff as the minimal necessary duration of input processing required to uphold predictive reliability. 
Thus, we define optimal cutoff timestep (OCT) as the smallest timestep $ \hat{t} \in\{1,2...,T\} $ at which the SNN prediction function \( f(\boldsymbol{X}, t) \) remain correct prediction for all future timesteps \( t \) greater than \( \hat{t} \), formulated as
\begin{equation} \label{eq:con1}
\text{OCT}(\boldsymbol{X}) = \min \{ \hat{t} \mid \forall t > \hat{t}, f(\boldsymbol{X}, t) \text{ is correct}\}.
\end{equation} 
This equation expresses the earliest timestep from which the prediction function $f(\cdot)$ can confidently and correctly classify according to the partial input. The function $\text{OCT}(\boldsymbol{X})$ represents a theoretical lower bound of average inference timestep, ensuring that each input sample can undergo minimal inference timestep without sacrificing accuracy. 

In our evaluations, we will utilise the OCT as a new metric to assess the performance of SNN models in terms of their cutoff efficiency. This metric will enable us to quantitatively determine the lower boundary of the cutoff for any given sample, pinpointing the earliest timestep at which our SNN models can deliver reliable predictions. 

\subsubsection{Top-K Gap for Cutoff Approximation}\label{sec:gap-based}
During runtime inference, a critical challenge emerges due to the unpredictability of future predictions, which makes determining the OCT in real time impractical. To address this, we introduce a cutoff mechanism based on the gap between the largest (top-1) and second largest (top-2) output logits. This `Top-K' approach suggests that a larger gap between these two output logits indicates a low likelihood of changing the prediction during inference, thereby marking an appropriate point for cutoff.

To formalise this concept, let $\text{Top}_k(\boldsymbol{Y}(t))$ be the top-$k$ logit of one neuron at the output layer, We define the function $Y_{\text{gap}}$ to represent the gap of top-1 and top-2 values of output $\boldsymbol{Y}(t)$ as

\begin{equation}
    Y_{\text{gap}}= \text{Top}_1(\boldsymbol{Y}(t)) - \text{Top}_2(\boldsymbol{Y}(t)).
\end{equation}

Then, we let $ D\{\cdot\}$  denote the inputs in subset of $D$ that satisfy a certain condition.  
Now, we can define the confidence rate $C(t, D\{Y_{\text{gap}}>\beta_t\})$ as 

\begin{equation} \label{eq:con2}
\begin{split}
&\textit{Confidence rate: } C(t, D\{Y_{\text{gap}}>\beta_t\}) = \\
&\frac{1}{|D\{Y_{\text{gap}}>\beta_t\}|}\sum_{\boldsymbol{X}\in D\{Y_{\text{gap}}>\beta_t\}} (\text{OCT}(\boldsymbol{X}) \leq   t).
\end{split}
\end{equation}
The confidence rate intuitively computes the percentage of inputs in $D$ that can achieve the prediction success at or before $T$.
$|D\{Y_{\text{gap}}>\beta_t\}|$ denotes the number of samples in $D$ satisfying the condition. It is not hard to see that, when $t=0$, $C(t, D\{Y_{\text{gap}}>\beta_t\})$ is also $0$, and with the increase of time $t$, $C(t, D\{Y_{\text{gap}}>\beta_t\})$ will also increase until reaching $1$.  
Our algorithm searches for a minimum $\beta_t \in \mathbb{R^+}$  at a specific $t$, as expressed in the following optimisation objective
\begin{equation} \label{eq:con3}
\arg\min_{\beta_t} C( t, D\{Y_{\text{gap}} > \beta_t\}) \geq 1-\epsilon,
\end{equation}
where $\epsilon$ is a pre-specified constant such that $1-\epsilon$ represents an acceptable level of confidence for activating cutoff. 

\begin{figure*}[t!]
	\includegraphics[width=0.98\textwidth]{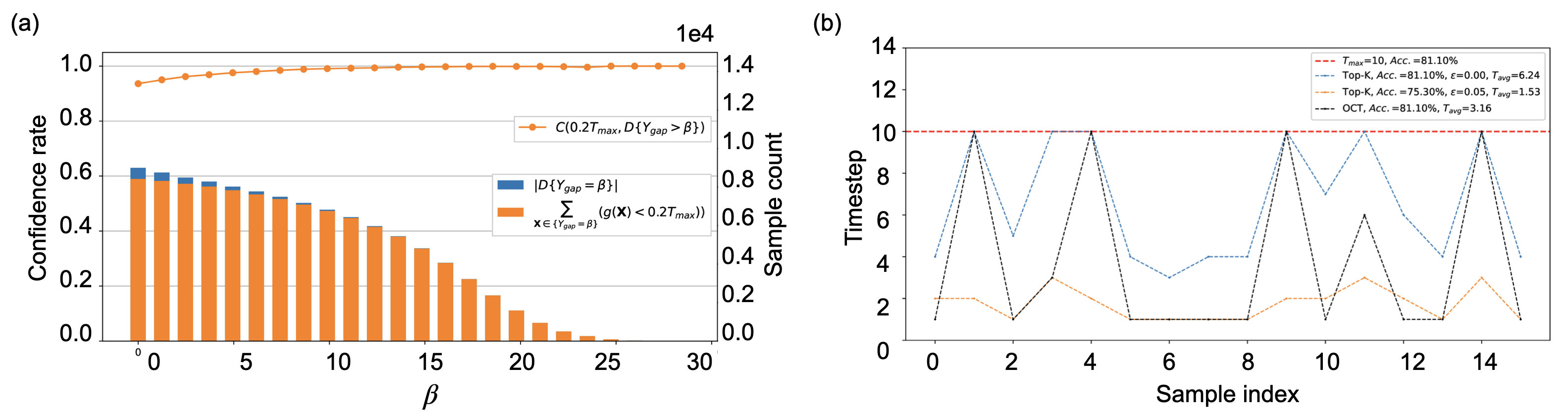}
    \caption{Evaluation of the Top-K cutoff on CIFAR10-DVS, using a directly trained SNN model with $T = 10$ (as detailed in Section \ref{sec:exp}): (a) The increase of $\beta$ limits the number of samples eligible for cutoff, while concurrently enhancing the confidence in the cutoff decision. (b) To enhance the readability, the inference timestep of 16 samples from the test dataset under varied trigger conditions. 
    }
    \label{fig:con_cov}
\end{figure*}

Figure \ref{fig:con_cov} provides a visual representation of equation \eqref{eq:con1} to \eqref{eq:con3}, illustrating the theoretical concepts comprehensively. Additionally, Figure \ref{fig:con_cov}(b) highlights the performance disparity between OCT and Top-K cutoff. 
The OCT identifies the minimum theoretical timestep $\hat t$ for each sample, ensuring predictions remain accurate as if processed until the maximum timestep $T$. Adjusting the $\epsilon$ parameter in the Top-K cutoff allows for reduced average timestep but may lead to a compromise in accuracy. This insight directs our primary optimisation goal towards enhancing SNN cutoff performance, a significant departure from traditional SNN optimisations \cite{bu2022optimal, deng2022temporal, yao2021temporal, fang2021deep, fang2021incorporating} that typically emphasise inference over a fixed duration.

\subsection{Optimising SNN for Cutoff}\label{sec:reg}

To improve the cutoff performance, we concentrate on maximising the cosine similarity between the actual spiking rate at time \( t \), \( \boldsymbol{r}^l(t) \), and the desired spiking rate, \( \boldsymbol{\tilde r}^l \). This objective is achieved by minimising the inverse of the cosine similarity between these two rates across hidden layers during training, formalised through the introduction of the regulariser of cosine similarity (RCS), defined as 

\begin{equation} \label{eq:cs_min}
\min_{\boldsymbol{W}} \left( \frac{\boldsymbol{\tilde r}^l \cdot \boldsymbol{r}^l(t)}{\Vert \boldsymbol{\tilde r}^l \Vert_2 \Vert \boldsymbol{r}^l(t)\Vert_2} \right)^{-1},
\end{equation}
where \( \boldsymbol{\tilde r}^l \) is desired spiking rate, and \( \boldsymbol{r}^l(t) \) denotes the spiking rate at time \( t \). Both spiking rates will be approximated differently according to conversion (Section \ref{sec:reg-conv}) and direct training (Section \ref{sec:reg-dt}). 

The motivation for using cosine similarity lies in its proven correlation with the final accuracy of quantised neural networks, as demonstrated by \cite{banner2018scalable}. In a similar vein, we hypothesize that a higher cosine similarity between $\boldsymbol{\tilde r}^l$ and  $\boldsymbol{r}^l(t)$ will correlate with a smaller accuracy drop at time $t$. However, approximating these spiking rates poses a significant challenge, particularly within the frameworks of ANN-to-SNN conversion and direct training.

To address these challenges, we differentiate the spiking rates in the context of conversion and direct training as follows \( \boldsymbol{\tilde r}_c^l \) and \( \boldsymbol{r}_c^l(t) \) denote the desired spiking rate and the actual spiking rate at time $t$, respectively, in the conversion-based method with IF neurons, while \( \boldsymbol{\tilde r}_d^l \) and  \( \boldsymbol{r}_d^l(t) \) are used for direct training with LIF neurons. Each method necessitates a distinct approach to approximate these rates, reflecting their unique operational contexts. 

It is important to note that there is no established evidence that equation \eqref{eq:cs_min} directly optimises accuracy at maximum timestep $T$. 
This is because the cosine similarity term primarily serves as a penalty for the alignment between features at different timesteps rather than an explicit measure of distance to the ground truth.
Therefore, to avoid any potential degradation in the original training performance, our RCS is applied selectively, focusing only on those samples that yield correct predictions at $T$.

\subsubsection{Regularising ANN before conversion} \label{sec:reg-conv}
For conversion-based SNN, \cite{wu2022little} introduced on a fundamental relationship between spiking rates  $r_c^l(t)$ and ReLU activation $\boldsymbol{a}^l$, which gives

\begin{equation} \label{eq:spiking-rate}
 \boldsymbol{r}_c^l(t) = \frac{1}{V^l_{\text{thr}}} \Big(\boldsymbol{W}^l\boldsymbol{r}_c^{l-1}(t)+\boldsymbol{b}^l \Big)- \boldsymbol{\Delta}^l(t),
\end{equation}
where \( \boldsymbol{r}_c^l(t) = \frac{1}{t} \sum^t \boldsymbol{\theta}^l(t') \) denotes the spiking rate at time \( t \), 
with \( t' \) representing each discrete timestep leading up to \( t \) at the at $l$-th layer, and
$\boldsymbol{\Delta}^l(t) \triangleq  {\boldsymbol{V}^{l+}(t)}/{(t V^l_{\text{thr}})}$
represents the residual spiking rate. 

The spiking rate at the first layer can be initialised as $\boldsymbol{r}_c^1(t) = \boldsymbol{a}^1/V^1_{thr} - \boldsymbol{\Delta}^1(t)$. When \( t \) is sufficiently large to make \( \boldsymbol{\Delta}^l(t)\) negligible, we have the desired spiking rate for the \( l \)-th layer as 
\begin{equation} \label{eq:rd}
\boldsymbol{\tilde r}_c^l = \frac{\boldsymbol{a}^l}{V^1_{thr}}.
\end{equation}

Givens that $\boldsymbol{r}_c^l(t) = \boldsymbol{\tilde r}_c^l - \boldsymbol{\Delta}^l(t)$, the cosine similarity is given by

\begin{equation} \label{eq:ineq}
\begin{split}
\frac{\boldsymbol{\tilde r}_c^l\cdot\boldsymbol{r}_c^l(t)}{\Vert \boldsymbol{\tilde r}_c^l \Vert_2 \Vert \boldsymbol{r}_c^l(t)\Vert_2} 
& \geq \frac{\Vert  \boldsymbol{\tilde r}_c^l\Vert_2}{\Vert  \boldsymbol{\tilde r}_c^l \Vert_2 + \Vert \boldsymbol{\Delta}^l(t) \Vert_2} \\
& = \frac{\Vert  \boldsymbol{a}^l \Vert_2}{\Vert  \boldsymbol{a}^l \Vert_2 + \Vert \boldsymbol{V}^{l+}(t)/(t) \Vert_2}. \\
\end{split}
\end{equation}

Assuming that elements in $\boldsymbol{V}^{l+}(t)$ satisfy uniform distribution over the time $t$ and they are in $[0,V_{\text{thr}}]$, we can derive boundary for expected value of $\Vert \boldsymbol{V}^{l+}(t)/t \Vert_2$ as $\mathbb{E}(\Vert \boldsymbol{V}^{l+}(t)/t \Vert_2)\leq \sqrt{n^l}V_{\text{thr}
}/(\sqrt{3}t)$ (proof in Supplementary Material). Moreover, at high dimensions, the relative error made as considering $\mathbb{E}(\Vert \boldsymbol{V}^{l+}(t)/t \Vert_2)$ instead of the random variable $\Vert \boldsymbol{V}^{l+}(t)/t \Vert_2$ becomes asymptotically negligible \cite{biau2015high}. Therefore, the lower bound of equation \eqref{eq:ineq} is given by
\begin{equation} \label{eq:lowerbound}
\begin{split}
\frac{\boldsymbol{\tilde r}_c^l\cdot\boldsymbol{r}_c^l(t)}{\Vert \boldsymbol{\tilde r}_c^l \Vert_2 \Vert \boldsymbol{r}_c^l(t)\Vert_2} 
&\geq \frac{\Vert \boldsymbol{a}^l \Vert_2}{\Vert \boldsymbol{a}^l \Vert_2 + \sqrt{n^l} V_{\text{thr}}^l/(\sqrt{3}t)} \\
&= \frac{\sqrt{3}t}{\sqrt{3}t+\sqrt{n^l}V_{\text{thr}}^l/\Vert \boldsymbol{a}^l \Vert_2}.
\end{split}
\end{equation} 
This equation explicitly explains that: (a) the increase of $t$ to $t \gg \sqrt{n^l} V_{\text{thr}}^l/\Vert \boldsymbol{a}^l \Vert_2$, where $n^l$ is the number of elements in activation $\boldsymbol{a}^l$, can increase the lower bound, and (b) it is possible to  minimise term $\sqrt{n^l}V_{\text{thr}}^l/\Vert \boldsymbol{a}^l \Vert_2$ for developing an SNN with optimised performance at any time during the inference. In other words, for a conversion-based SNN to achieve optimal cutoff performance, the model expects a good (small) ratio of threshold voltage $V_{\text{thr}}^l$ to average accumulated current, i.e., $\Vert \boldsymbol{a}^l \Vert_2/\sqrt{n^l}$, while not degrading SNN classification performance. 

\begin{algorithm}[t]
\caption{Compute RCS loss in ANN training} \label{alg:conversion}
\begin{algorithmic}[1]
\REQUIRE Dataset $\mathcal{D}$, ANN prediction function $f_{ann}(\cdot)$, $L$ total layer number, batch size $B$
\FOR{each batch $\mathcal{B}$ in $\mathcal{D}$ of size $B$}
    \STATE Initialize $\boldsymbol{A}^l_{\text{batch}}$ to store activations for layer $l$ 
    \STATE Initialize $\boldsymbol{A}^l_{norm}$ to store L2 norms for layer $l$ 
    \FOR{each data point $\boldsymbol{\bar X}$ in $\mathcal{B}$}
        \IF{$f_{ann}(\boldsymbol{\bar X})$ is correct} 
            \STATE Append layer $l$ activations $\boldsymbol{a}^l$ to $\boldsymbol{A}^l_{\text{batch}}$
        \ENDIF
    \ENDFOR
    \STATE Compute $A_{max}$ as the maximum value in $\boldsymbol{A}^l_{\text{batch}}$
    \FOR{each $\boldsymbol{a}^l$ in $\boldsymbol{A}^l_{\text{batch}}$}
        \STATE Append $\Vert \boldsymbol{a}^l \Vert_2$ to $\boldsymbol{A}^l_{norm}$
    \ENDFOR
    \STATE Compute $\boldsymbol{A}^l_{min}$ as the minimum value in $\boldsymbol{A}^l_{norm}$
    \STATE $L_{RCS} \leftarrow \frac{1}{L}\sum_l^L \sqrt{n^l}\frac{A^l_{max}}{\boldsymbol{A}^l_{min}}$
\ENDFOR
\end{algorithmic}
\end{algorithm}
In ANN training, we aim to affect the training process to result in SNN for optimal cutoff. For this, Algorithm \ref{alg:conversion} has been designed to increase the lower bound defined in equation \eqref{eq:lowerbound}. Following \cite{rueckauer2017conversion, wu2022little}, we use the maximum activation value, denoted as $A_{max}$ in the algorithm, to approximate the threshold voltage $V^l_{\text{thr}}$ at each layer. The algorithm optimises activation ratios within each layer during training, directly enhancing SNN performance.

\subsubsection{Regularising Direct Training} \label{sec:reg-dt}
In the case of direct training, SNN indicates its potential at processing spatio-temporal data, leading to a dynamic spiking rate \(\boldsymbol{r}_d^l(t)\) throughout the inference process. This capability, highlighted in \cite{yao2021temporal}, sets direct training apart from ANN-to-SNN conversion, which is designed only for static input. Given that spikes can vary across timesteps in direct training, and their average may not accurately capture the temporal information, we approximate the spiking rate at each timestep by normalising the membrane potential at that specific timestep. Thus, $\boldsymbol{r}_d^l(t)$ is defined as

\begin{equation}
    \boldsymbol{r}_d^l(t) = \boldsymbol{\theta}^l(t) + \frac{\boldsymbol{V}^{l+}(t)}{V^l_{\text{thr}}},
\end{equation}
where $\boldsymbol{\theta}^l(t)$  is generated spikes and the normalised residual membrane potential $\boldsymbol{V}^{l+}(t)/V_{thr}$ reflects the firing intention of neurons.

Moving on to the next step, we focus on computing the desired spiking rate $\tilde{\boldsymbol{r}}_d$. The trend that SNN accuracy improves with an increased number of the timestep, attributed to the accumulation of relevant information over time, is illustrated by Figure \ref{fig:general_cutoff} in Section \ref{sec:exp}. This observation forms the basis for our estimation of $\tilde{\boldsymbol{r}}_d$, which is calculated as 

\begin{equation} \label{eq: dt_rd}
    \tilde{\boldsymbol{r}}^l_d = \frac{1}{N} \sum^{T}_{t=T-N} \boldsymbol{r}_d^l(t),
\end{equation}
where $N$ is an integer hyperparameter. This method posits that the later spiking rates could serve as more representative information for a given sample. As indicated in equation \eqref{eq:con1} from Section \ref{sec:cutoff-theory}, consistent and correct predictions are crucial for optimal cutoff performance. Thus, we only optimise $\boldsymbol{r}_d(t)$ using equation \eqref{eq:cs_min} only when the last $N$ predictions are correct. The details of this approach are described in Algorithm \ref{alg:dt}.

\begin{algorithm}
\caption{Compute RCS loss in SNN training} \label{alg:dt}
\begin{algorithmic}[1]
\REQUIRE Dataset $\mathcal{D}$, SNN prediction function $f(\cdot)$, $L$ total layer number, batch size $B$
\FOR{each batch $\mathcal{B}$ in $\mathcal{D}$ of size $B$}
    \STATE Initialize $\boldsymbol{S}^l_{\text{batch}}$ to store cosine similarity for the layer $l$
    \FOR{each data point $\boldsymbol{X}$ in $\mathcal{B}$}
        \STATE Set $consistent\_correct = \text{True}$
        \FOR{$t = T - N$ to $T$}
            \IF{$f(\boldsymbol{X},t)$ is not correct}
                \STATE Set $consistent\_correct = \text{False}$
            \ENDIF
        \ENDFOR
        
        \IF{$consistent\_correct$ is True}
            \FOR{each layer $l$}
                \STATE Compute $\tilde{\boldsymbol{r}}_d = \frac{1}{N} \sum^{T}_{t=T-N} \boldsymbol{r}_d^l(t)$
                \STATE Compute $\frac{1}{T} \sum_t^{T} \frac{\tilde{\boldsymbol{r}}_d^l \cdot \boldsymbol{r}_d^l(t)}{\Vert \tilde{\boldsymbol{r}}_d^l \Vert_2 \Vert \boldsymbol{r}_d^l(t)\Vert_2}$  \\ and append to $\boldsymbol{S}^l_{\text{batch}}$  
            \ENDFOR
        \ENDIF
    \ENDFOR
    \FOR{each layer $l$}
        \STATE Compute $\boldsymbol{S}^l_{\text{min}}$ as the minimum value in $\boldsymbol{S}^l_{\text{batch}}$
        \STATE $L_{RCS} \leftarrow \frac{1}{L}\sum_l^L (\boldsymbol{S}^l_{\text{min}})^{-1}$
    \ENDFOR
\ENDFOR
\end{algorithmic}
\end{algorithm}

\subsubsection{Optimisation Objective} \label{sec:reg-obj}

Building on the $L_{RCS}$  computed in Algorithms \ref{alg:conversion} and \ref{alg:dt} for ANN and SNN, respectively, we integrate the regularisation term into our overall optimisation objective for each model. $L_{RCS}$ captures the worst-case features across layers during training, and minimising $L_{RCS}$ helps mitigate these issues. We follow \cite{deng2022temporal,bu2022optimal} to use cross-entropy loss (denoted as $L_{ce}$) as the primary training loss. The final training objective with the RCS regularisation is
\begin{equation}
\min_{\boldsymbol{W}} \left( L_{ce} + \alpha L_{RCS} \right),
\end{equation}
where $\boldsymbol{W}$ is the weight matrix of ANN or SNN to be trained, and $\alpha$ is a hyper-parameter to balance two loss terms.

\section{Experiment}\label{sec:exp}
In this section, we engage in comprehensive experiments to evaluate SNN models using our newly proposed metric `OCT', in conjunction with the `Top-K' cutoff and the `RCS' regularisation technique. These experiments are designed to explore the compatibility of our approaches with prevalent SNN training methods. For example, in the conversion-based training, we implement the quantised clip-floor-shift (QCFS) method from \cite{bu2022optimal}. This method replaces ReLU with the QCFS activation function to reduce the accuracy loss after conversion. For direct training, we adopt the temporal efficient training (TET)  \cite{deng2022temporal} and Temporal efficient batch normalisation (TEBN) \cite{duan2022temporal}, both representing the most recent developments in SNN training algorithms. For clarity, each configuration in our experimental setup is denoted within brackets: QCFS($\cdot$) indicates the quantisation length, while TET($\cdot$) and TEBN($\cdot$) refer to the maximum timestep for training. Addtionally, Top-K($\cdot$) indicates the setting of $\epsilon$. 
For an easy reference to the incorporated techniques in the models, we use notations such as `TET($\cdot$) + RCS, w/ Top-K', indicating that the SNN model has been enhanced with both RCS regularisation and the Top-K cutoff strategy. In our experiments, the ANN-to-SNN conversion method, such as QCFS, is applied only to frame-based datasets. We do not evaluate it on event-based datasets, which inherently involve temporal dynamics, as it primarily focuses on spatial information. 

\subsection{Experimental Datasets and Setup}\label{exp:dataset}

The experiments are conducted across a variety of datasets, including both frame-based and event-based inputs, and network architectures. Specifically, we evaluate our approaches across diverse settings: ResNet-18 \cite{he2016deep} for CIFAR10/100 \cite{krizhevsky2009learning} and Tiny-ImageNet \cite{le2015tiny}, VGGSNN \cite{deng2022temporal} for CIFAR10-DVS \cite{li2017cifar10dvs} and N-Caltech101 \cite{orchard2015converting}, along with a 5-layer convolutional network \cite{fang2021incorporating} for DVS128 Gesture \cite{amir2017low}. To effectively process event-based datasets, we implement a downsampling strategy by integrating a $4\times4$ kernel with a stride of 4 at the beginning of the original network architecture. This adjustment can directly feed event data into the SNN, as suggested by \cite{shrestha2018slayer}.
\begin{figure*}[!ht]
    \includegraphics[width=0.98\textwidth]{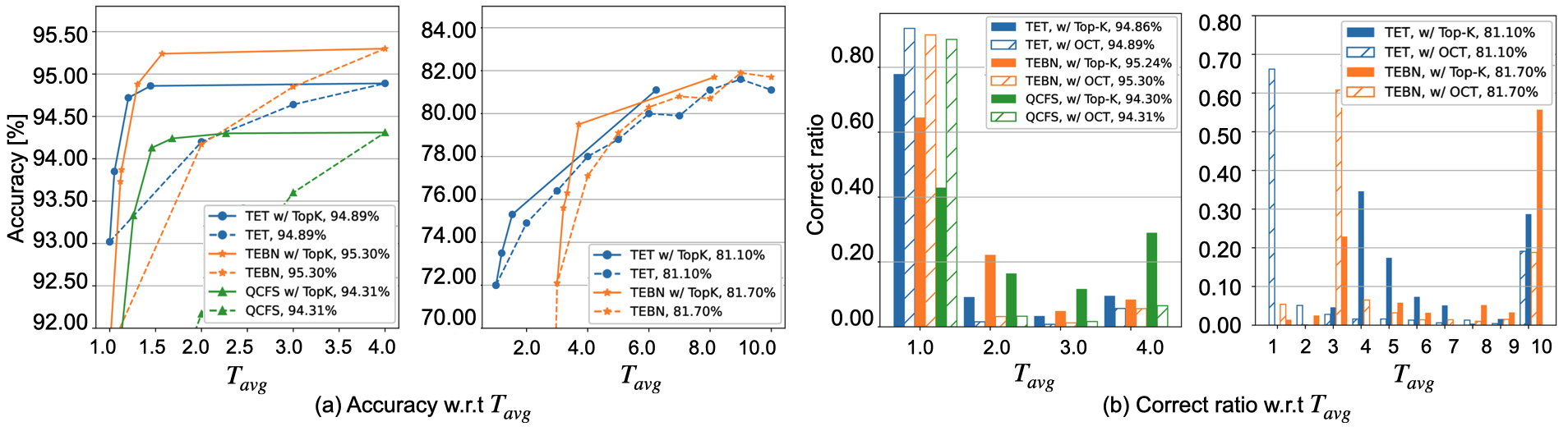}

	\caption{Comparison of SNN with and without Top-K cutoff on CIFAR10 (left) and CIFAR10-DVS (right) across various training methods: (a) The Top-K cutoff is determined by $\epsilon$ values ranging from 0.00 to 0.50 in increments of 0.05. (b) The statistic data is extracted from testing samples under $\epsilon=0.02$ for CIFAR10 $\epsilon=0.0$ for CIFAR10-DVS.}
	\label{fig:general_cutoff}
\end{figure*}
The samples in the event-based datasets record the event addresses with on/off events over a specific duration. For CIFAR10-DVS, it consists of 10,000 samples extracted from CIFAR10 \cite{li2017cifar10dvs}. Each sample has $128\times128$ spatial resolution. The length of each spike train is less than or equal to $1.3s$. For N-Caltech101, it has 8709 samples categorised into 101 classes. The number of samples in each class ranges from 31 to 800. The length of each spike train is about $0.3s$. The width in the x-direction does not exceed 240 pixels, and in the y-direction does not exceed 180 pixels. For these two datasets, we use 90\% samples in each class for training and 10\% for testing. DVS128 Gesture consists of 1341 samples with 11 categories. Each sample is repetitive over $6.0s$. Due to the repetitive information in these samples, we select the first $1.3s$. All event-based samples are split into 10 frames for the training and evaluation.

For training, the hyper-parameter $\alpha$ varied depending on the training method and dataset, chosen from {0.001, 0.002, 0.003} for conversion-based methods and {0.001, 0.003, 0.005} for event-based datasets. For frame-based datasets like CIFAR10/100 and Tiny-ImageNet, we used a batch size of 128 and 300 epochs. The auto augmentation \cite{cubuk2019autoaugment} is deployed for Cifar-10/100 for accuracy enhancement. For event-based datasets, the training parameters were set to 100 epochs with batch sizes of 128 for CIFAR10-DVS, 64 for N-Caltech101, and 32 for DVS128 Gesture.

Evaluation of the Top-K cutoff requires a set of $\beta$ values derived from the training dataset. To evaluate the efficiency, each sample is simulated until its corresponding OCT $ \hat{t} $. We use $ T_{\text{avg}} $ to represent the average number of timesteps required for the inputs from the entire test dataset. Since models often exhibit overconfidence post-training, we integrated dropout layers with a 0.3 dropout rate after each spiking layer to counteract this effect during characterisation. The efficacy of this dropout is explained in \cite{srivastava2014dropout,jin2022weight}. To implement RCS in direct training, Equation~\eqref{eq: dt_rd} suggests that the later spiking rates are expected to align with the desired spiking rates. Thus, we set $N = 1$ for $T = 4$ and $N = 3$ for $T = 10$, guided by the ratio $N/T \approx 0.3$.

\subsection{Experimental Results} \label{exp:accexp}

Figure \ref{fig:general_cutoff}(a) demonstrates the implementation of the Top-K cutoff across various SNN training methods, showcasing its ability to enhance computational efficiency by reducing the number of timesteps required to achieve comparable accuracy. Specifically, the SNN requires \textbf{1.76 to 2.76$\times$ fewer timesteps} for CIFAR-10 with a near-zero accuracy drop of 0.01\% to 0.06\%, and \textbf{1.23 to 1.60$\times$ fewer timesteps} for CIFAR10-DVS remaining same accuracy. However, while Top-K serves as a practical approximation, a discernible gap still exists between this empirical approach and the theoretical cutoff by OCT. As illustrated in Figure \ref{fig:general_cutoff}(b), the Top-K cutoff effectively implements an adaptive timestep strategy but falls short of achieving the ideal cutoff, particularly in terms of correctly classifying samples at the first timestep compared to OCT. To provide further insight, we revisit Figure \ref{fig:con_cov}(b) from Section \ref{sec:gap-based}, which visually illustrates the impact of $\epsilon$ on Top-K cutoff across different samples. A larger $\epsilon$ may indicate fewer timestep for each sample, but the accuracy drops. This highlights the need to optimise the Top-K cutoff during training via RCS.

\begin{table}[!ht]
    \caption{Comparison of accuracy and latency before and after applying RCS across different training methods on frame-based datasets. OCT reflects the theoretical cutoff performance, indicating undiminished accuracy and the average optimal cutoff timesteps. Top-K cutoff with $\epsilon$ values of 0.0 and 0.05 demonstrates the performance of the approximated cutoff.} 
    \label{tab:framecomparison}
    \begin{subtable}{0.50\textwidth}
    \centering
    \small
    \begin{tabular*}{\linewidth}{@{\extracolsep{\fill}}  c c c c c c c}
    \Xhline{4\arrayrulewidth}
         &\multicolumn{2}{c}{OCT} &\multicolumn{2}{c}{Top-K(0)} &\multicolumn{2}{c}{Top-K(0.05)}\\    
    \hline
     Method &Acc. &$T_{avg}$  &Acc. &$T_{avg}$ &Acc. &$T_{avg}$ \\
    \Xhline{4\arrayrulewidth}           
      QCFS(4)   & 94.04 &  1.27  & 94.04 & 4.00 & 94.04 &2.40\\ 
      QCFS(4) + \textbf{RCS}  & \textbf{94.32} &  \textbf{1.26}  &94.32 & 4.00  & 94.32 &  2.94\\ 
      \hline
      TET(4)   & 94.89 &  1.20  & 94.89 & 4.00 & 94.86& 1.45 \\ 
      TET(4) +  \textbf{RCS}  & \textbf{95.24} &  \textbf{1.19} &95.24 & 2.10  &95.23 & 1.67 \\ 
      \hline
      TEBN(4)   & 95.30 &  1.22  & 95.30 & 4.00 & 95.24 & 1.57 \\ 
      TEBN(4) +  \textbf{RCS}  & \textbf{95.49}  &  1.22  & 95.49 & 4.00 & 95.47 & 1.58 \\ 
    \Xhline{4\arrayrulewidth}           
    \end{tabular*}
    \caption{CIFAR10.}
    \end{subtable}
    \begin{subtable}{0.50\textwidth}
    \centering
    \small
    \begin{tabular*}{\linewidth}{@{\extracolsep{\fill}}  c c c c c c c}
    \Xhline{4\arrayrulewidth}
         &\multicolumn{2}{c}{OCT} &\multicolumn{2}{c}{Top-K(0)} &\multicolumn{2}{c}{Top-K(0.05)}\\    
    \hline
     Method &Acc. &$T_{avg}$  &Acc. &$T_{avg}$ &Acc. &$T_{avg}$ \\
    \Xhline{4\arrayrulewidth}             

      QCFS(4)   & 75.20 &  2.03  & 75.20 & 4.00 & 75.20 & 3.80\\ 
   
      QCFS(4) + \textbf{RCS}   & \textbf{76.21} & \textbf{2.00}  & 76.21 &  4.00  & 76.21 &  3.72\\ 

      \hline
      TET(4)     & 77.02 &  1.84  & 77.02 & 4.00 &77.02 & 3.30 \\  
      TET(4) + \textbf{RCS}   & \textbf{77.81} &  \textbf{1.81}  & 77.81 &  4.00 & 77.81 &  3.23 \\ 
      
      \hline
      TEBN(4)        & 77.93 &  1.85  & 77.93 & 4.00 & 77.93 & 3.03 \\    
      TEBN(4) + \textbf{RCS}  & \textbf{78.13} &  1.85  & 78.13 &  4.00 & 78.13 &  3.08 \\ 
    
    \Xhline{4\arrayrulewidth}           
    \end{tabular*}
    \caption{CIFAR100.}\label{tab:table-cifar100}
    \end{subtable}
     \begin{subtable}{0.50\textwidth}
    \centering
    \small
    \begin{tabular*}{\linewidth}{@{\extracolsep{\fill}}  c c c c c c c}
    \Xhline{4\arrayrulewidth}
         &\multicolumn{2}{c}{OCT} &\multicolumn{2}{c}{Top-K(0)} &\multicolumn{2}{c}{Top-K(0.05)}\\    
    \hline
     Method &Acc. &$T_{avg}$  &Acc. &$T_{avg}$ &Acc. &$T_{avg}$ \\
    \Xhline{4\arrayrulewidth}             

      QCFS(4)   & 47.11 &  2.97  & 47.11 & 3.98 & 47.11 & 3.91\\ 
   
      QCFS(4) + \textbf{RCS}   & \textbf{47.71} & \textbf{2.94}  & 47.71 &  3.99  & 47.71 &  3.97\\ 

      \hline
      TET(4)     & 56.56 &  2.52  & 56.56 & 3.74 & 56.56 & 3.65\\ 
      TET(4) + \textbf{RCS}   & \textbf{56.69} &  2.52 & 56.69 &  3.75  & 56.69 &  3.56\\  
      
      \hline
      TEBN(4)        & 56.19 &  2.56  & 56.19 & 3.87 &56.19 & 3.63 \\   
      TEBN(4) + \textbf{RCS}  & \textbf{56.69} &  \textbf{2.54}  & 56.69 &  3.82 & 56.69 &  3.51 \\ 
    
    \Xhline{4\arrayrulewidth}           
    \end{tabular*}
    \caption{Tiny-ImageNet.}
    \end{subtable}
       
\end{table}

\begin{table}[!ht]
    \caption{Comparison of accuracy and latency before and after applying RCS across direct training methods on event-based datasets.}    \label{tab:eventcomparison}
    \begin{subtable}{0.50\textwidth}
    \centering
    \small
    \begin{tabular*}{\linewidth}{@{\extracolsep{\fill}}  c c c c c c c}
    \Xhline{4\arrayrulewidth}
         &\multicolumn{2}{c}{OCT} &\multicolumn{2}{c}{Top-K(0)} &\multicolumn{2}{c}{Top-K(0.05)}\\    
    \hline
     Method &Acc. &$T_{avg}$  &Acc. &$T_{avg}$ &Acc. &$T_{avg}$ \\
    \Xhline{4\arrayrulewidth}           

      TET(10)   & 81.10 &  3.16  & 81.10 & 6.24 & 75.30 &1.53 \\ 
      TET(10) + \textbf{RCS}  & \textbf{83.10} &  \textbf{3.05} &  83.10 & 6.10 &76.90 &1.43\\ 
      \hline
      TEBN(10)     & 81.70 &  4.46  & 81.70 & 8.14  & 79.95 & 3.71\\ 
      TEBN(10) + \textbf{RCS}  & \textbf{82.20} &  \textbf{3.94} & 82.20 & 7.88 & 79.10 & 3.27 \\ 
    \Xhline{4\arrayrulewidth}           
    \end{tabular*}
    \caption{CIFAR10-DVS.}
    \end{subtable}
    \begin{subtable}{0.50\textwidth}
    \centering
    \small
    \begin{tabular*}{\linewidth}{@{\extracolsep{\fill}}  c c c c c c c}
    \Xhline{4\arrayrulewidth}
         &\multicolumn{2}{c}{OCT} &\multicolumn{2}{c}{Top-K(0)} &\multicolumn{2}{c}{Top-K(0.05)}\\    
    \hline
     Method &Acc. &$T_{avg}$  &Acc. &$T_{avg}$ &Acc. &$T_{avg}$ \\
    \Xhline{4\arrayrulewidth}           

      TET(10)   & 85.01 &2.64 &85.01 & 7.39 &83.48&1.96 \\ 
      TET(10) + \textbf{RCS}  & \textbf{85.66} &  \textbf{2.57} & 85.67 & 6.06 &84.03 &1.77 \\ 
      \hline
      TEBN(10)     & 82.49 &  3.01  & 82.49 &  7.44   & 79.32 & 1.86 \\ 
      TEBN(10) + \textbf{RCS}  & \textbf{83.15} &  \textbf{2.99}  & 83.15 & 5.95 & 80.20 & 1.91 \\ 
    \Xhline{4\arrayrulewidth}           
    \end{tabular*}
    \caption{N-Caltrech101.}
    \end{subtable}
    \begin{subtable}{0.50\textwidth}
    \centering
    \small
    \begin{tabular*}{\linewidth}{@{\extracolsep{\fill}}  c c c c c c c}
    \Xhline{4\arrayrulewidth}
         &\multicolumn{2}{c}{OCT} &\multicolumn{2}{c}{Top-K(0)} &\multicolumn{2}{c}{Top-K(0.05)}\\    
    \hline
     Method &Acc. &$T_{avg}$  &Acc. &$T_{avg}$ &Acc. &$T_{avg}$ \\
    \Xhline{4\arrayrulewidth}           

      TET(10)   & 96.97 &  2.18  & 96.97 & 6.31 & 96.21 & 4.28 \\     
      TET(10) + \textbf{RCS}  & \textbf{97.35}&  \textbf{1.66} &  97.35 & 5.14 & 95.83 & 2.53 \\ 
      \hline
      TEBN(10)    & 96.21  &  3.00  & 96.21 & 9.3 & 96.21 & 7.42\\     
      TEBN(10) + \textbf{RCS}  & \textbf{96.97} &  \textbf{2.73}  & 96,97 & 9.1 &96.97& 7.21\\       
    \Xhline{4\arrayrulewidth}           
    \end{tabular*}
    \caption{DVS128 Gesture.}
    \end{subtable}
\end{table}

\begin{figure*}[!ht]
	\includegraphics[width=\textwidth]{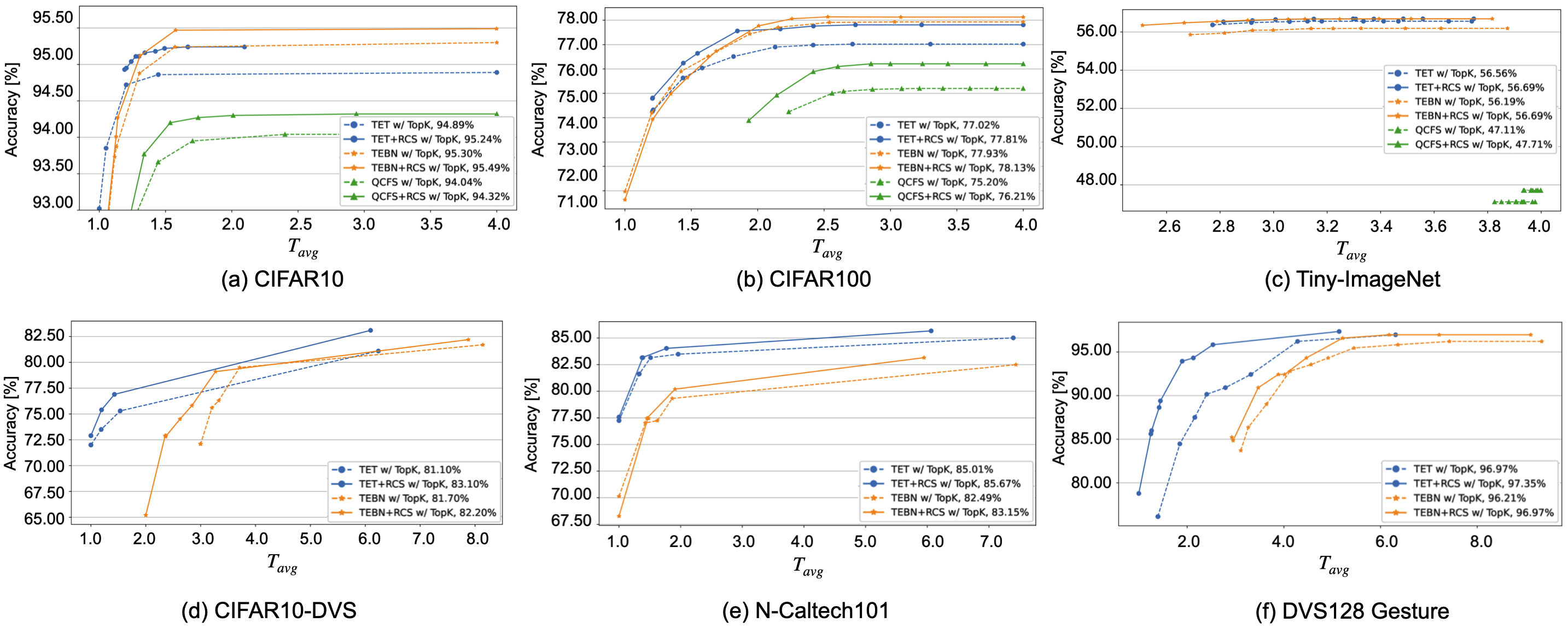}

	\caption{Comparison of Top-K cutoff accuracy before (dashed lines) and after (solid lines) regularisation, across a range of $\epsilon$ values from 0.00 to 0.50, increasing in steps of 0.05. The accuracy for full-length input is detailed in the legend.}
	\label{fig:acc-topk}
\end{figure*}

Given that RCS is designed to complement the OCT, our experimental results consider the OCT as a key metric associated with Top-K on frame-based inputs (Table \ref{tab:framecomparison}) and event-based inputs (Table \ref{tab:eventcomparison}). As seen in both tables, a smaller OCT always indicates better accuracy performance and a more effective cutoff. 

In Table \ref{tab:framecomparison}, RCS facilitates a significant OCT reduction for the QCFS models, ranging from 0.01 to 0.03 across CIFAR10/100 and Tiny-ImageNet. Specifically, on CIFAR100, this improvement is marked by a 0.3 decrease in OCT alongside a 1.01\% boost in accuracy, underlining the effectiveness of RCS in enhancing SNN from ANN training.
Conversely, when RCS is applied to direct training methods like TET and TEBN, the improvements on frame-based datasets appear more modest. This limited improvement may be due to direct training methods are designed to optimise the network’s performance at small timesteps, such as configuring a small maximum timestep ($T=4$) during training.
Furthermore, OCT estimates the theoretically optimal cutoff point (without accuracy loss), reflecting the upper bound of accuracy achievable during cutoff. In practical, the Top-K cutoff, as an approximation of OCT, achieves accuracy less than or equal to that of OCT, depending on the setting of $\epsilon$.

The effect of RCS becomes more significant for event-based datasets, as shown in Table \ref{tab:eventcomparison}, with OCT reductions ranging from 0.02 to 0.52. Notably, for the TET with RCS, Top-K(0) enables SNN to achieve zero accuracy loss across all event-based datasets while requiring \textbf{1.64 to 1.95$\times$ fewer timesteps}. This performance surpasses that of TET without RCS, which requires only \textbf{1.35 to 1.60$\times$ fewer timesteps}. Moreover, Figure \ref{fig:acc-topk} illustrates that the implementation of RCS shifts the accuracy curve upward compared to the curve without RCS, which means that the similar accuracy can have less inference timesteps.

\subsection{Comparison with Existing Research on Cutoff}
Previous studies have investigated adaptive inference strategies in SNNs to improve inference efficiency. For instance, SEENN \cite{li2023seenn} employs reinforcement learning to train an auxiliary network jointly with the SNN, enabling early exits, while the dynamic confidence (DC) strategy \cite{li2023unleashing} relies on post-training calibration.

\begin{figure*}[!ht]
	\includegraphics[width=\textwidth]{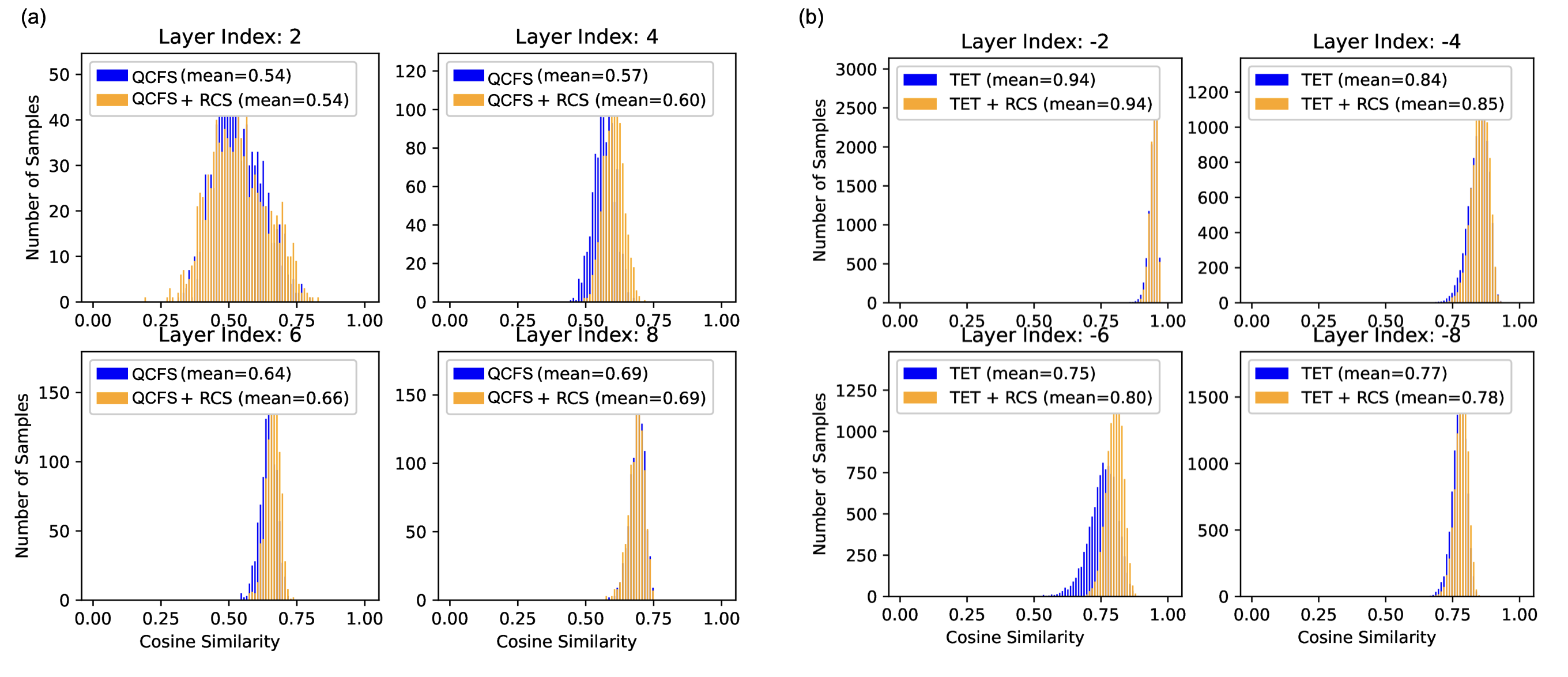}
	\caption{Histograms of cosine similarity at different layers for SNNs trained with and without the RCS regulariser. (a) Results for SNN (ResNet-18) trained with QCFS on CIFAR10 at $t=2$. (b) Results for SNN (VGGSNN) trained with TET on CIFAR10-DVS at $t=3$. Each subplot corresponds to a specific layer index. Note that layer index $-2$ indicates the the second last layer.}\label{fig:hist_cs}
\end{figure*}

In contrast, our approach eliminates the need for auxiliary networks, offering a more efficient cutoff solution, and leverages the RCS to assist training, enhancing cutoff robustness. Figure~\ref{fig:hist_cs} demonstrates the impact of the RCS on feature alignment. The histograms show that RCS consistently improves cosine similarity across layers, indicating better alignment between early timestep features and the expected features. Table \ref{tab:comparision_exiting_work} compares our results with SNN models that employ adaptive inference strategies . The results indicate that integrating RCS and Top-K SNN cutoff techniques leads to superior performance at low timestep.

\begin{table*}[!ht]
    \caption{Comparison of accuracy and latency between our methods and state-of-the-art SNN work. The $\epsilon$ value for each model is adjusted within the range [0, 0.5] with step size of 0.01 to evaluate accuracy at specific timesteps.}
    \label{tab:comparision_exiting_work}
    \centering
    \small
    \renewcommand{\arraystretch}{1.2} 
    \begin{tabular*}{\linewidth}{@{\extracolsep{\fill}} c|c|c|c|c|c|}
    \Xhline{4\arrayrulewidth}
    \makecell{\textbf{Dataset} \\ \textbf{(Architecture)}} & \makecell{\textbf{Training} \\ \textbf{Framework}} & \makecell{\textbf{Method}} & \makecell{\textbf{Acc. ($T_{avg,1}$)}} & \makecell{\textbf{Acc. ($T_{avg,2}$)}} & \makecell{\textbf{Acc. ($T_{avg,3}$)}} \\
    \Xhline{4\arrayrulewidth}
    \multirow{5}{*}{\shortstack{CIFAR10 \\ (ResNet-18)}} 
        & CFFS \cite{li2023unleashing} 
        & DC & 94.11 (2.52) & - & - \\ \cline{2-6}
        & \multirow{3}{*}{QCFS}
        & DC & 94.27 (11.51) & - & - \\ 
        
        & &SEENN-I & 95.08 (2.01) & 93.63 (1.40) & 91.08 (1.18) \\ 
        & & RCS+Top-K & 94.31 (1.71) & 93.96 (1.46) & 91.59 (1.12) \\ \cline{2-6}
        & \multirow{1}{*}{TET}
        &  \textbf{RCS+Top-K} & \textbf{95.23 (1.58)} & \textbf{95.16 (1.32)} & \textbf{94.38 (1.07)} \\
    \Xhline{4\arrayrulewidth}
    \multirow{3}{*}{\shortstack{CIFAR100 \\ (ResNet-18)}} 

        &\multirow{2}{*}{QCFS}  & SEENN-I & 65.48 (6.19) & 56.99 (4.41) & 39.33 (2.57) \\ 
        & & RCS+Top-K & 76.21 (3.33) & 75.22 (2.23) & 71.94 (1.75) \\  \cline{2-6}
        & \multirow{1}{*}{TET} 
        & RCS+Top-K & \textbf{77.81 (2.73)} & \textbf{77.57 (1.91)} & \textbf{76.17 (1.45)} \\ \cline{2-6}
    \Xhline{4\arrayrulewidth}
    \multirow{3}{*}{\shortstack{CIFAR10-DVS \\ (VGGSNN)}} 
        & \multirow{3}{*}{TET}
        & SEENN-I & 82.7 (5.17) & 77.60 (2.53) & - \\ 
        & & SEENN-II  & 82.6 (4.49) & - & - \\
        & & RCS+Top-K & \textbf{82.80 (4.14)} & \textbf{81.10 (2.52)} & \textbf{76.90 (1.43)} \\
    \Xhline{4\arrayrulewidth}
    \end{tabular*}
    
\end{table*}

\section{Conclusions}
In this paper, we focus on developing SNN that achieve high efficiency throughout both training and inference processes, making them particularly well-suited for inferring with adaptive timestep. We introduce two key innovations designed to enhance SNN performance: a regulariser targets the training phase, and a cutoff mechanism optimises the inference stage. Our comprehensive experiments showcase these advancements, marking notable enhancements in accuracy and inference efficiency over traditional methods. The Top-K cutoff technique introduced here proves to be versatile across various SNN neuron models, such as IF and LIF, provided the predictions rely solely on output analysis. 

While our approach shows strong potential, some limitations remain. The current evaluation is primarily conducted on VGG and ResNet architectures, and its applicability to advanced architectures like spiking transformers is yet to be explored. Additionally, the regularizer increases memory requirements, as seen with a 26\% rise in memory usage—from 8.9 GB to 11.3 GB—when directly training an SNN ($T=4$) with ResNet-18 on CIFAR10. This overhead could pose challenges for training larger-scale networks. Furthermore, our evaluation relies on GPU-based simulations rather than hardware implementations.

Future work could explore integrating these techniques with advanced architectures, such as spiking transformers. Expanding the evaluation to more complex datasets and real-world applications would provide a deeper understanding of the methods' scalability and practical utility. Further, optimising the RCS to reduce its memory overhead would improve its suitability for larger-scale models, addressing resource constraints. Finally, implementing and testing the cutoff with the SNN on Field Programmable Gate Arrays (FPGAs) would provide valuable insights into its performance and feasibility for real-world applications.

\bibliographystyle{IEEEtran} 
\bibliography{main}

\begin{thebibliography}{10}
\providecommand{\url}[1]{#1}
\csname url@rmstyle\endcsname
\providecommand{\newblock}{\relax}
\providecommand{\bibinfo}[2]{#2}
\providecommand\BIBentrySTDinterwordspacing{\spaceskip=0pt\relax}
\providecommand\BIBentryALTinterwordstretchfactor{4}
\providecommand\BIBentryALTinterwordspacing{\spaceskip=\fontdimen2\font plus
\BIBentryALTinterwordstretchfactor\fontdimen3\font minus \fontdimen4\font\relax}
\providecommand\BIBforeignlanguage[2]{{%
\expandafter\ifx\csname l@#1\endcsname\relax
\typeout{** WARNING: IEEEtran.bst: No hyphenation pattern has been}%
\typeout{** loaded for the language `#1'. Using the pattern for}%
\typeout{** the default language instead.}%
\else
\language=\csname l@#1\endcsname
\fi
#2}}

\bibitem{davies2018loihi}
M.~Davies, N.~Srinivasa, T.-H. Lin, G.~Chinya, Y.~Cao, S.~H. Choday, G.~Dimou, P.~Joshi, N.~Imam, S.~Jain, \emph{et~al.}, ``Loihi: A neuromorphic manycore processor with on-chip learning,'' \emph{IEEE Micro}, vol.~38, no.~1, pp. 82--99, 2018.

\bibitem{akopyan2015truenorth}
F.~Akopyan, J.~Sawada, A.~Cassidy, R.~Alvarez-Icaza, J.~Arthur, P.~Merolla, N.~Imam, Y.~Nakamura, P.~Datta, G.-J. Nam, \emph{et~al.}, ``Truenorth: Design and tool flow of a 65 mw 1 million neuron programmable neurosynaptic chip,'' \emph{IEEE transactions on computer-aided design of integrated circuits and systems}, vol.~34, no.~10, pp. 1537--1557, 2015.

\bibitem{DVS:2008}
P.~Lichtsteiner, C.~Posch, and T.~Delbruck, ``A 128$\times $128 120 db 15$\mu$s latency asynchronous temporal contrast vision sensor,'' \emph{IEEE journal of solid-state circuits}, vol.~43, no.~2, pp. 566--576, 2008.

\bibitem{DVS:2010}
T.~Delbr{\"u}ck, B.~Linares-Barranco, E.~Culurciello, and C.~Posch, ``Activity-driven, event-based vision sensors,'' in \emph{Proceedings of 2010 IEEE International Symposium on Circuits and Systems}, pp. 2426--2429, 2010.

\bibitem{DVS:2020}
G.~Gallego, T.~Delbr{\"u}ck, G.~Orchard, C.~Bartolozzi, B.~Taba, A.~Censi, S.~Leutenegger, A.~J. Davison, J.~Conradt, K.~Daniilidis, \emph{et~al.}, ``Event-based vision: A survey,'' \emph{IEEE transactions on pattern analysis and machine intelligence}, vol.~44, no.~1, pp. 154--180, 2020.

\bibitem{DAS:2018}
J.~Anumula, D.~Neil, T.~Delbruck, and S.-C. Liu, ``Feature representations for neuromorphic audio spike streams,'' \emph{Frontiers in Neuroscience}, vol.~12, p.~23, 2018.

\bibitem{amir2017low}
A.~Amir, B.~Taba, D.~Berg, T.~Melano, J.~McKinstry, C.~Di~Nolfo, T.~Nayak, A.~Andreopoulos, G.~Garreau, M.~Mendoza, \emph{et~al.}, ``A low power, fully event-based gesture recognition system,'' in \emph{Proceedings of the IEEE Conference on Computer Vision and Pattern Recognition}, pp. 7243--7252, 2017.

\bibitem{kim2021n}
J.~Kim, J.~Bae, G.~Park, D.~Zhang, and Y.~M. Kim, ``N-imagenet: Towards robust, fine-grained object recognition with event cameras,'' in \emph{Proceedings of the IEEE/CVF International Conference on Computer Vision}, pp. 2146--2156, 2021.

\bibitem{messikommer2020event}
N.~Messikommer, D.~Gehrig, A.~Loquercio, and D.~Scaramuzza, ``Event-based asynchronous sparse convolutional networks,'' in \emph{European Conference on Computer Vision}.\hskip 1em plus 0.5em minus 0.4em\relax Springer, pp. 415--431, 2020.

\bibitem{rueckauer2017conversion}
B.~Rueckauer, I.-A. Lungu, Y.~Hu, M.~Pfeiffer, and S.-C. Liu, ``Conversion of continuous-valued deep networks to efficient event-driven networks for image classification,'' \emph{Frontiers in Neuroscience}, vol.~11, p. 682, 2017.

\bibitem{sengupta2019going}
A.~Sengupta, Y.~Ye, R.~Wang, C.~Liu, and K.~Roy, ``Going deeper in spiking neural networks: {VGG} and residual architectures,'' \emph{Frontiers in Neuroscience}, vol.~13, p.~95, 2019.

\bibitem{deng2021optimal}
S.~Deng and S.~Gu, ``Optimal conversion of conventional artificial neural networks to spiking neural networks,'' in \emph{International Conference on Learning Representations}, 2021.

\bibitem{bu2022optimal}
T.~Bu, W.~Fang, J.~Ding, P.~DAI, Z.~Yu, and T.~Huang, ``Optimal {ANN}-{SNN} conversion for high-accuracy and ultra-low-latency spiking neural networks,'' in \emph{International Conference on Learning Representations}, 2022.

\bibitem{han2020rmp}
B.~Han, G.~Srinivasan, and K.~Roy, ``{RMP-SNN}: Residual membrane potential neuron for enabling deeper high-accuracy and low-latency spiking neural network,'' in \emph{Proceedings of the IEEE/CVF Conference on Computer Vision and Pattern Recognition}, pp. 13\,558--13\,567, 2020.

\bibitem{wu2018spatio}
Y.~Wu, L.~Deng, G.~Li, J.~Zhu, and L.~Shi, ``Spatio-temporal backpropagation for training high-performance spiking neural networks,'' \emph{Frontiers in Neuroscience}, vol.~12, p. 331, 2018.

\bibitem{wu2019direct}
Y.~Wu, L.~Deng, G.~Li, J.~Zhu, Y.~Xie, and L.~Shi, ``Direct training for spiking neural networks: Faster, larger, better,'' in \emph{Proceedings of the AAAI conference on artificial intelligence}, vol.~33, pp. 1311--1318, 2019.

\bibitem{fang2021deep}
W.~Fang, Z.~Yu, Y.~Chen, T.~Huang, T.~Masquelier, and Y.~Tian, ``Deep residual learning in spiking neural networks,'' \emph{Advances in Neural Information Processing Systems}, vol.~34, pp. 21\,056--21\,069, 2021.

\bibitem{fang2021incorporating}
W.~Fang, Z.~Yu, Y.~Chen, T.~Masquelier, T.~Huang, and Y.~Tian, ``Incorporating learnable membrane time constant to enhance learning of spiking neural networks,'' in \emph{Proceedings of the IEEE/CVF International Conference on Computer Vision}, pp. 2661--2671, 2021.

\bibitem{diehl2015fast}
P.~U. Diehl, D.~Neil, J.~Binas, M.~Cook, S.-C. Liu, and M.~Pfeiffer, ``Fast-classifying, high-accuracy spiking deep networks through weight and threshold balancing,'' in \emph{2015 International joint conference on neural networks (IJCNN)}.\hskip 1em plus 0.5em minus 0.4em\relax ieee, pp. 1--8, 2015.

\bibitem{wu2022little}
D.~Wu, X.~Yi, and X.~Huang, ``A little energy goes a long way: Build an energy-efficient, accurate spiking neural network from convolutional neural network,'' \emph{Frontiers in Neuroscience}, vol.~16, 2022.

\bibitem{pmlr-v139-li21d}
Y.~Li, S.~Deng, X.~Dong, R.~Gong, and S.~Gu, ``A free lunch from ann: Towards efficient, accurate spiking neural networks calibration,'' in \emph{Proceedings of the 38th International Conference on Machine Learning}, ser. Proceedings of Machine Learning Research, vol. 139, pp. 6316--6325, 2021.

\bibitem{Rathi2020Enabling}
N.~Rathi, G.~Srinivasan, P.~Panda, and K.~Roy, ``Enabling deep spiking neural networks with hybrid conversion and spike timing dependent backpropagation,'' in \emph{International Conference on Learning Representations}, 2020.

\bibitem{rathi2021diet}
N.~Rathi and K.~Roy, ``{DIET-SNN}: A low-latency spiking neural network with direct input encoding and leakage and threshold optimization,'' \emph{IEEE Transactions on Neural Networks and Learning Systems}, 2021.

\bibitem{wu2021progressive}
J.~Wu, C.~Xu, X.~Han, D.~Zhou, M.~Zhang, H.~Li, and K.~C. Tan, ``Progressive tandem learning for pattern recognition with deep spiking neural networks,'' \emph{IEEE Transactions on Pattern Analysis and Machine Intelligence}, 2021.

\bibitem{yao2021temporal}
M.~Yao, H.~Gao, G.~Zhao, D.~Wang, Y.~Lin, Z.~Yang, and G.~Li, ``Temporal-wise attention spiking neural networks for event streams classification,'' in \emph{Proceedings of the IEEE/CVF International Conference on Computer Vision}, pp. 10\,221--10\,230, 2021.

\bibitem{neftci2019surrogate}
E.~O. Neftci, H.~Mostafa, and F.~Zenke, ``Surrogate gradient learning in spiking neural networks: Bringing the power of gradient-based optimization to spiking neural networks,'' \emph{IEEE Signal Processing Magazine}, vol.~36, no.~6, pp. 51--63, 2019.

\bibitem{li2021differentiable}
Y.~Li, Y.~Guo, S.~Zhang, S.~Deng, Y.~Hai, and S.~Gu, ``Differentiable spike: Rethinking gradient-descent for training spiking neural networks,'' \emph{Advances in Neural Information Processing Systems}, vol.~34, pp. 23\,426--23\,439, 2021.

\bibitem{kim2021revisiting}
Y.~Kim and P.~Panda, ``Revisiting batch normalization for training low-latency deep spiking neural networks from scratch,'' \emph{Frontiers in Neuroscience}, vol.~15, p. 773954, 2021.

\bibitem{duan2022temporal}
C.~Duan, J.~Ding, S.~Chen, Z.~Yu, and T.~Huang, ``Temporal effective batch normalization in spiking neural networks,'' in \emph{Advances in Neural Information Processing Systems}, 2022.

\bibitem{guo2023membrane}
Y.~Guo, Y.~Zhang, Y.~Chen, W.~Peng, X.~Liu, L.~Zhang, X.~Huang, and Z.~Ma, ``Membrane potential batch normalization for spiking neural networks,'' in \emph{Proceedings of the IEEE/CVF International Conference on Computer Vision}, pp. 19\,420--19\,430, 2023.

\bibitem{guo2023joint}
Y.~Guo, W.~Peng, Y.~Chen, L.~Zhang, X.~Liu, X.~Huang, and Z.~Ma, ``Joint a-snn: Joint training of artificial and spiking neural networks via self-distillation and weight factorization,'' \emph{Pattern Recognition}, vol. 142, p. 109639, 2023.

\bibitem{zhang2024enhancing}
Y.~Zhang, X.~Liu, Y.~Chen, W.~Peng, Y.~Guo, X.~Huang, and Z.~Ma, ``Enhancing representation of spiking neural networks via similarity-sensitive contrastive learning,'' in \emph{Proceedings of the AAAI Conference on Artificial Intelligence}, vol.~38, no.~15, pp. 16\,926--16\,934, 2024.

\bibitem{zhou2023spikformer}
Z.~Zhou, Y.~Zhu, C.~He, Y.~Wang, S.~YAN, Y.~Tian, and L.~Yuan, ``Spikformer: When spiking neural network meets transformer,'' in \emph{The Eleventh International Conference on Learning Representations}, 2023.

\bibitem{zhang2022spiking}
J.~Zhang, B.~Dong, H.~Zhang, J.~Ding, F.~Heide, B.~Yin, and X.~Yang, ``Spiking transformers for event-based single object tracking,'' in \emph{Proceedings of the IEEE/CVF conference on Computer Vision and Pattern Recognition}, pp. 8801--8810, 2022.

\bibitem{wang2023masked}
Z.~Wang, Y.~Fang, J.~Cao, Q.~Zhang, Z.~Wang, and R.~Xu, ``Masked spiking transformer,'' in \emph{Proceedings of the IEEE/CVF International Conference on Computer Vision}, pp. 1761--1771, 2023.

\bibitem{guo2024ternary}
Y.~Guo, Y.~Chen, X.~Liu, W.~Peng, Y.~Zhang, X.~Huang, and Z.~Ma, ``Ternary spike: Learning ternary spikes for spiking neural networks,'' in \emph{Proceedings of the AAAI Conference on Artificial Intelligence}, vol.~38, no.~11, pp. 12\,244--12\,252, 2024.

\bibitem{ding2024shrinking}
Y.~Ding, L.~Zuo, M.~Jing, P.~He, and Y.~Xiao, ``Shrinking your timestep: Towards low-latency neuromorphic object recognition with spiking neural networks,'' in \emph{Proceedings of the AAAI Conference on Artificial Intelligence}, vol.~38, no.~10, pp. 11\,811--11\,819, 2024.

\bibitem{li2023seenn}
Y.~Li, T.~Geller, Y.~Kim, and P.~Panda, ``{SEENN}: Towards temporal spiking early-exit neural networks,'' \emph{arXiv preprint arXiv:2304.01230}, 2023.

\bibitem{li2023unleashing}
C.~Li, E.~Jones, and S.~Furber, ``Unleashing the potential of spiking neural networks by dynamic confidence,'' \emph{arXiv preprint arXiv:2303.10276}, 2023.

\bibitem{chen2023agreeing}
J.~Chen, S.~Park, and O.~Simeone, ``Agreeing to stop: Reliable latency-adaptive decision making via ensembles of spiking neural networks,'' \emph{arXiv preprint arXiv:2310.16675}, 2023.

\bibitem{shafer2008tutorial}
G.~Shafer and V.~Vovk, ``A tutorial on conformal prediction.'' \emph{Journal of Machine Learning Research}, vol.~9, no.~3, 2008.

\bibitem{ruckauer2019closing}
B.~R{\"u}ckauer, N.~K{\"a}nzig, S.-C. Liu, T.~Delbruck, and Y.~Sandamirskaya, ``Closing the accuracy gap in an event-based visual recognition task,'' \emph{arXiv preprint arXiv:1906.08859}, 2019.

\bibitem{deng2022temporal}
S.~Deng, Y.~Li, S.~Zhang, and S.~Gu, ``Temporal efficient training of spiking neural network via gradient re-weighting,'' in \emph{International Conference on Learning Representations}, 2022.

\bibitem{banner2018scalable}
R.~Banner, I.~Hubara, E.~Hoffer, and D.~Soudry, ``Scalable methods for 8-bit training of neural networks,'' \emph{Advances in Neural Information Processing Systems}, vol.~31, 2018.

\bibitem{biau2015high}
G.~Biau and D.~M. Mason, ``High-dimensional $p$ $p$-norms,'' in \emph{Mathematical statistics and limit theorems}.\hskip 1em plus 0.5em minus 0.4em\relax Springer, pp. 21--40, 2015.

\bibitem{he2016deep}
K.~He, X.~Zhang, S.~Ren, and J.~Sun, ``Deep residual learning for image recognition,'' in \emph{Proceedings of the IEEE Conference on Computer Vision and Pattern Recognition}, pp. 770--778, 2016.

\bibitem{krizhevsky2009learning}
A.~Krizhevsky, G.~Hinton, \emph{et~al.}, ``Learning multiple layers of features from tiny images,'' \emph{Toronto, ON, Canada}, 2009.

\bibitem{le2015tiny}
Y.~Le and X.~Yang, ``Tiny imagenet visual recognition challenge,'' \emph{CS 231N}, vol.~7, no.~7, p.~3, 2015.

\bibitem{li2017cifar10dvs}
H.~Li, H.~Liu, X.~Ji, G.~Li, and L.~Shi, ``Cifar10-{DVS}: an event-stream dataset for object classification,'' \emph{Frontiers in Neuroscience}, vol.~11, p. 309, 2017.

\bibitem{orchard2015converting}
G.~Orchard, A.~Jayawant, G.~K. Cohen, and N.~Thakor, ``Converting static image datasets to spiking neuromorphic datasets using saccades,'' \emph{Frontiers in Neuroscience}, vol.~9, p. 437, 2015.

\bibitem{shrestha2018slayer}
S.~B. Shrestha and G.~Orchard, ``Slayer: Spike layer error reassignment in time,'' \emph{Advances in Neural Information Processing Systems}, vol.~31, 2018.

\bibitem{cubuk2019autoaugment}
E.~D. Cubuk, B.~Zoph, D.~Mane, V.~Vasudevan, and Q.~V. Le, ``Autoaugment: Learning augmentation strategies from data,'' in \emph{Proceedings of the IEEE/CVF Conference on Computer Vision and Pattern Recognition}, pp. 113--123, 2019.

\bibitem{srivastava2014dropout}
N.~Srivastava, G.~Hinton, A.~Krizhevsky, I.~Sutskever, and R.~Salakhutdinov, ``Dropout: a simple way to prevent neural networks from overfitting,'' \emph{The journal of machine learning research}, vol.~15, no.~1, pp. 1929--1958, 2014.

\bibitem{jin2022weight}
G.~Jin, X.~Yi, P.~Yang, L.~Zhang, S.~Schewe, and X.~Huang, ``Weight expansion: A new perspective on dropout and generalization,'' \emph{Transactions on Machine Learning Research}, 2022.

\end{thebibliography}
\appendix
\renewcommand{\thetable}{A-\arabic{table}}
\renewcommand{\thefigure}{A-\arabic{figure}}

\setcounter{figure}{0}    
\setcounter{equation}{0}
\setcounter{table}{0}
\appendices

\section{Inequation proof}\label{app:proof}

We follow \cite{banner2018scalable} to derive the bound of expected norm of a random variable vector. By Jensen's inequality, it gives
\begin{equation} 
\begin{split}
\mathbb{E}(\Vert \boldsymbol{V}(t)\Vert_2) = \mathbb{E}(\sqrt{\sum_i V_i(t)^2}) 
&\leq \sqrt{\mathbb{E}(\sum_i V_i(t)^2)} \\
&= \sqrt{\sum_i \mathbb{E}(V_i(t)^2)}
\end{split}
\end{equation}
As the $V_i(t)$ is a uniform random variable in range $[0,V_{\text{thr}}]$, the expected value of $V_i^2(t)$  can be computed as follows
\begin{equation}
\mathbb{E}(V(t)_i^2) = \int_{0}^{V_{\text{thr}}} x^2\frac{1}{V_{\text{thr}}} dx = \frac{V_{\text{thr}}^2}{3}
\end{equation}
which yields
\begin{equation} \label{eq:a-3}
\mathbb{E}(\Vert \boldsymbol{V}(t)\Vert_2) \leq \frac{\sqrt{n}V_{\text{thr}}}{\sqrt{3}}
\end{equation}
Since $t$ is constant value, the following inequality holds
\begin{equation} 
\mathbb{E}(\Vert \boldsymbol{V}(t)/t\Vert_2) \leq \frac{\sqrt{n}V_{\text{thr}}}{\sqrt{3}t}
\end{equation}

\end{document}


\title{Supplementary: Optimising Event-Driven Spiking Neural Network with \\ Regularisation and Cutoff}

\author{Dengyu Wu\\
Institution1\\
Institution1 address\\
{\tt\small firstauthor@i1.org}
\and
Second Author\\
Institution2\\
First line of institution2 address\\
{\tt\small secondauthor@i2.org}
}

\maketitle

\renewcommand{\thetable}{S-\arabic{table}}
\renewcommand{\thefigure}{S-\arabic{figure}}

\setcounter{figure}{0}    
\setcounter{equation}{0}
\setcounter{table}{0}
\section{Inequation proof}\label{app:proof}
We follow \cite{banner2018scalable} to derive the bound of expected norm of a random variable vector. By Jensen's inequality, it gives
\begin{equation} 
\begin{split}
\mathbb{E}(\Vert \boldsymbol{V}(t)\Vert_2) = \mathbb{E}(\sqrt{\sum_i V_i(t)^2}) &\leq \sqrt{\mathbb{E}(\sum_i V_i(t)^2)} \\
&= \sqrt{\sum_i \mathbb{E}(V_i(t)^2)} 
\end{split}
\end{equation}
As the $V_i(t)$ is a uniform random variable in range $[0,V_{thr}]$, the expected value of $V_i^2(t)$  can be computed as follows
\begin{equation}
\mathbb{E}(V(t)_i^2) = \int_{0}^{V_{thr}} x^2\frac{1}{V_{thr}} dx = \frac{V_{thr}^2}{3}
\end{equation}
which yields
\begin{equation} \label{eq:a-3}
\mathbb{E}(\Vert \boldsymbol{V}(t)\Vert_2) \leq \frac{\sqrt{n}V_{thr}}{\sqrt{3}}
\end{equation}
Since $t$ and $S_r$ are constant values, the following inequality holds
\begin{equation} 
\mathbb{E}(\Vert \boldsymbol{V}(t)/(tS_r)\Vert_2) \leq \frac{\sqrt{n}V_{thr}}{\sqrt{3}tS_r}
\end{equation}
\section{Experiment setup}\label{app:experiment}
The network architectures for difference datasets are given in Table \ref{tab:architecture}, which are modified from VGG-11 \cite{VGG:2014} for Cifar10-DVS \& N-Caltech101 and VGG-like structure \cite{fang2021incorporating} for DVS128 Gesture. C64k8s4 represents the convolutional layer with $filters=64$, $kernel~size=4$ and $strides=4$.  The default values of Kernel size and strides are 3 and 1 respectively. AP2 is the average pooling layer, MP2 is the max pooling layer with $kernel~size=2$ and FC is the fully-connected layer. 

\begin{table}[h!]
    \centering
    \footnotesize
    \caption{Network architectures for difference datasets.}
    \begin{tabular*}{0.48\textwidth} { c c}
     \Xhline{4\arrayrulewidth}
     Dataset  & Network Architecture\\
     \Xhline{4\arrayrulewidth}
     Cifar10-DVS  & C64k8s4-C64-C128-C256s2-C256-C512s2\\
      N-Caltech101 &-C512-C512s2-C512-AP2-FC512-Output \\
     \hline
     DVS128 Gesture & C128k8s4-\{C128-MP2\}*5
     -FC512-FC128-Output \\
     \Xhline{4\arrayrulewidth}             
    \end{tabular*}
    \label{tab:architecture}
\end{table}

Batch Normalisation \cite{pmlr-v37-ioffe15} is applied after each convolutional and fully-connected layer to accelerate the convergence of ANN training. For all experiment, the learning rate is set to 0.1 and decays to zero after 300 epochs based on cosine decay schedule \cite{loshchilov2016sgdr}. Weight decay is set to 0.0005. We set $\alpha$ to 0.003 for the regulariser proposed in Section 4.2 and use pixel shifting as the data augmentation for all models, i.e., both width and height are randomly shifted by the range [-20\%,20\%]. Dropout is applied after fully-connected layer for DVS128 Gesture to improve the training and the dropout rate is 0.2. For conversion method, we set the batch size to 128 for $F=1$ and 32 for $F>1$ to reduce memory consumption. Furthermore, we use the same architecture for the SNN-LIF model, except for SNN-LIF on DVS128 Gesture, where we replace max pooling with average pooling for better performance.

\subsection{Applying temporal training in ANN}\label{app:tt}

\begin{figure}[ht!]
	\includegraphics[width=0.48\textwidth]{Figures/temporal_training.eps}
	\caption{Forward propagation in temporal training.} 
	\label{fig:tt}
\end{figure}
Similar to the direct training in \cite{fang2021incorporating}, the temporal training in ANN, shown in Fig. \ref{fig:tt},  reshapes the temporal frames before forwarding them into the neural network and computes average loss after multiple outputs for optimisation. However, temporal training uses ReLU as the activation function and has no iterative operation during forward propagation. Although it ignores the correlation between the neighbouring frames in hidden layers, our experiment shows that SNN still can achieve good performance. Normally, iterative operation can be expensive when the number of iteration is large, i.g., large memory required \cite{fang2021incorporating}. Fig. \ref{fig:frames} presents that the increase of F can improve the accuracy on DVS128 Gesture, while it has little effect on Cifar10-DVS. We did not examine $F$ in N-Caltch101 due to its large size. 

\begin{figure}[h!]
    \centering
	\includegraphics[width=0.48\textwidth]{Figures/frames.eps}
	\caption{ANN accuracy w.r.t $F$ on DVS dataset.} 
	\label{fig:frames}
\end{figure}

Moreover, since temporal training treats consecutive frames as individual frames and generates most spike for the prediction, regulariser and cutoff can be directly deployed when $F>1$, 

\section{Additional Experiments}

\begin{figure}[!ht]
	\includegraphics[width=0.48\textwidth]{Figures/activation.eps}
    \caption{Comparison of Normalised activation distribution of ANN using different methods, ROE, COE and QA, for Cifar10-DVS. The dashed line indicates the $99.99th$ percentile of activation.}
    \label{app:activation}
\end{figure}
Fig. \ref{app:activation} presents the activation distributions for ROE, COE and QA over different layers. Comparing with QA, ROE helps ANN retain a Gaussian-like distribution. On the other hand, ROE brings the $99.99th$ percentile value closer to the mean of the activation than COE.
\begin{figure}[!ht]
    \centering
    \begin{subfigure}[t]{0.48\textwidth}
	\includegraphics[width=\textwidth]{Figures/batch_size.eps}
     \end{subfigure}
    \caption{Impact of $\alpha$ and batch size on resulted SNN at early inference time, for Cifar10-DVS. $method@batch~size$ denotes training method with the setting of batch size. 'ROE' represents $\alpha=0.003$ and 'Normal' means training without ROE. The accuracy at 0.500$T_{total}$ is shown in the bracket.}
    \label{app:batchsize}
\end{figure}

Fig. \ref{app:batchsize} indicates that the change of $\alpha$ and batch size can influence the performance of resulted SNN. It can be easily found that ROE-SNN can achieve better accuracy at $0.125T_{total}$ with larger batch size. However, too small or large batch size can significantly degrade the normal ANN training. Thus, we set the batch size wisely that can result in an optimal SNN.

Fig. \ref{fig:con_appendix} shows the confidence comparison on N-Caltech101 and DVS128 Gesture. The results are consistent, SNN-ROE can have better confidence at early inference time. 

Fig. \ref{fig:cs-roe} shows that the proposed regulariser improves $cos(\phi)$, i.e., the cosine similarity between spiking rate $r(t)$ and desired spiking rate $r_d$, over different layer and reduces the conversion error at early inference time. After increasing $\alpha$ above 0.003, the improvement in $cos(\phi)$ becomes limited. Therefore, we remain $\alpha = 0.003$ in other experiments. Fig. \ref{fig:cs-all} shows the difference of $cos(\phi)$ on SNN using ROE, QA and COE. SNN-ROE has general higher $cos(\phi)$ than the other methods. Meanwhile, Fig. \ref{fig:cs-roe} and \ref{fig:cs-all} also show that the increase of inference time benefits $cos(\phi)$.

\begin{figure*}[!ht]
    \begin{subfigure}[t]{0.48\textwidth}
	\includegraphics[width=\textwidth,height=0.48\textwidth]{Figures/confidence_ncaltech101.eps}
	\end{subfigure}
    \begin{subfigure}[t]{0.48\textwidth}
	\includegraphics[width=\textwidth,height=0.48\textwidth]{Figures/confidence_dvs_gesture.eps}
	\end{subfigure}
	\caption{Confidence w.r.t time ratio. for N-Caltech101 (left) and DVS128 Gesture (right).} 
	\label{fig:con_appendix}
\end{figure*}

\begin{figure*}[!ht]
    \begin{subfigure}[t]{0.48\textwidth}
	\includegraphics[width=\textwidth,height=0.48\textwidth]{Figures/cs_reg_3_8.eps}
	\end{subfigure}
    \begin{subfigure}[t]{0.48\textwidth}
	\includegraphics[width=\textwidth,height=0.48\textwidth]{Figures/cs_reg_4_8.eps}
	\end{subfigure}
    \caption{Comparison of $cos(\phi)$ for SNN-ROE w.r.t different setting of $\alpha$, at 0.375$T_{total}$ (left) and 0.500$T_{total}$ (right), on Cifar10-DVS. The conversion error is shown in the bracket.} 
	\label{fig:cs-roe}
\end{figure*}

\begin{figure*}[!ht]
    \begin{subfigure}[t]{0.48\textwidth}
	\includegraphics[width=\textwidth,height=0.48\textwidth]{Figures/cs_methods_3_8.eps}
	\end{subfigure}
    \begin{subfigure}[t]{0.48\textwidth}
	\includegraphics[width=\textwidth,height=0.48\textwidth]{Figures/cs_methods_4_8.eps}
	\end{subfigure}
    \caption{Comparison of $cos(\phi)$ for SNN-ROE ($\alpha = 0.003$), SNN-QA and SNN-COE, at 0.375$T_{total}$ (left) and 0.500$T_{total}$ (right), on Cifar10-DVS. The conversion error is shown in the bracket.} 
    \label{fig:cs-all}
\end{figure*}

\section{Notations}\label{app:notations}
We summarise the notations in Table \ref{tab:notations}.

\begin{table*}[t!]
    \caption{Notation summary.}  \label{tab:notations}

    \centering
    \begin{tabular}{ l l | l l }
     \Xhline{4\arrayrulewidth}
     Symbol &Definition &Symbol &Definition \\     
     \Xhline{4\arrayrulewidth}
      $i$ & Element index  & $l$ & Layer index\\    
      $T_{total}$ & Total inference time &$F$ & Number of input frame\\    
      $T$ & Duration of one frame & $t$ & Inference time \\   
      $N_{max}$ & Maximum spikes in training dataset & $S_r$ & Spiking resolution \\   
      $\boldsymbol{V}^l(t)$ &  Membrane potential& $\boldsymbol{Z}^l(t)$ & Weighted current \\
      $\boldsymbol{\theta}^l(t)$ &  Step function &$V_{thr}$ & Threshold voltage \\
      $\boldsymbol{W}^l$ &  Weight & $\boldsymbol{b}^l$ &  Bias \\    
      $\boldsymbol{X}(t)$ & Input spike train & $\hat{\boldsymbol{X}}_f$ & Spiking rate of f-th frame \\     
      $\boldsymbol{Y_f}$ & Output of $\hat{\boldsymbol{X}}_f$ & $\boldsymbol{Y}$ & Ground truth \\   
      $\boldsymbol{N}^l(t)$ &  Number of spikes received & $\boldsymbol{r}^l(t)$ & Spiking rate \\
      $\boldsymbol{\Delta}^l(t)$ &  Residual spiking rate & $\boldsymbol{\lambda}_d^l$ & Maximum value of activation  \\
      $\boldsymbol{r}_d^l$ & Desired spiking rate & ${\phi}^l$ &  Angular between $\boldsymbol{r}_d^l$ and $\boldsymbol{r}^l(t)$  \\
      $\boldsymbol{a}^l$ &  Activation  & $n$ & Dimension of $\boldsymbol{a}^l$ \\  
      $\boldsymbol{A}^l$ & A batch of activation & $L_{TT}$ & Temporal training loss \\
      $\hat{t}$ & Discrete inference time & $C(\hat{t},\cdot)$ & Confidence rate \\
      $g(\boldsymbol{X})$ & Gap of top-1 and top-2 number of spikes  & $\beta$ & Constant value for cutoff\\
      $D$ & Training dataset & $R(\cdot)$ & Regulariser term\\
      \Xhline{4\arrayrulewidth}
    \end{tabular}
\end{table*}

{\small
\bibliographystyle{ieee_fullname}
\bibliography{egbib}
}